\documentclass[pre,twocolumn,twoside,showpacs,superscriptaddress,floatfix]{revtex4-1}

\usepackage{graphics,amssymb,amsmath,epsf,color,hyperref}
\epsfclipon
\usepackage{epsfig,bm,epsf,graphics}
\usepackage{graphicx}% Include figure files
\usepackage{dcolumn}% Align table columns on decimal point
\usepackage{bm}% bold math
\usepackage{braket}% braket
\usepackage{color}
\usepackage{soul}
\usepackage{xcolor}
\usepackage{floatrow}
\usepackage{hyperref}

\hypersetup{colorlinks, citecolor=blue, linkcolor=blue}

\usepackage{graphicx}
\usepackage[normalem]{ulem}% added for \sout markup

\begin{document}

\title{Variational Garrote for Statistical Physics-based Sparse and Robust Variable Selection}

\author{Hyungjoon Soh}
\affiliation{Department of Physics Education, Seoul National University, Seoul 08826, Korea}

\author{Dongha Lee}
\affiliation{Department of Physics Education, Seoul National University, Seoul 08826, Korea}

\author{Vipul Periwal}
%\email[Corresponding author: ]{vipulp@mail.nih.gov}
\affiliation{Laboratory of Biological Modeling, National Institute of Diabetes and Digestive and Kidney Diseases, National Institutes of Health, Bethesda, Maryland 20892, USA}

\author{Junghyo Jo}
\email[Corresponding author: ]{jojunghyo@snu.ac.kr}
\affiliation{Department of Physics Education, Seoul National University, Seoul 08826, Korea}
\affiliation{Center for Theoretical Physics and Artificial Intelligence Institute, Seoul National University, Seoul 08826, Korea}
\affiliation{School of Computational Sciences, Korea Institute for Advanced Study, Seoul 02455, Korea}

\date{\today}

\begin{abstract}
Selecting key variables from high-dimensional data is increasingly important in the era of big data. Sparse regression serves as a powerful tool for this purpose by promoting model simplicity and explainability. In this work, we revisit a valuable yet underutilized method, the statistical physics-based Variational Garrote (VG), which introduces explicit feature selection spin variables and leverages variational inference to derive a tractable loss function. We enhance VG by incorporating modern automatic differentiation techniques, enabling scalable and efficient optimization.
We evaluate VG on both fully controllable synthetic datasets and complex real-world datasets. Our results demonstrate that VG performs especially well in highly sparse regimes, offering more consistent and robust variable selection than Ridge and LASSO regression across varying levels of sparsity. 
We also uncover a sharp transition: as superfluous variables are admitted, generalization degrades abruptly and the uncertainty of the selection variables increases.
This transition point provides a practical signal for estimating the correct number of relevant variables, an insight we successfully apply to identify key predictors in real-world data.
We expect that VG offers strong potential for sparse modeling across a wide range of applications, including compressed sensing and model pruning in machine learning.
\end{abstract}

%\pacs{05.45.Xt, 05.45.-a, 05.70.Fh}
%05.45.Xt: coupled oscillators, synchronization, nonlinear dynamics
%05.45.-a: dynamical systems, nonlinear dynamics
%05.70.Fh: phase transition in statistical mechanics and thermodynamics
 
\maketitle

%\tableofcontents

\section{Introduction}
Identifying relationships between variables is a fundamental task in science. Among various approaches, linear regression plays a central role in linking explanatory variables to dependent variables in statistical modeling~\cite{chatterjee2015regression, james2023linear}. Linear regression is useful in physics~\cite{majda2012physics, kaptanoglu2023sparse} for extracting equations of motion from time series data~\cite{Brunton2016} and for predicting trends in dynamical systems~\cite{Quade2016}, but its simplicity, interpretability, and predictive power make it a cornerstone of data analysis~\cite{maalouf2011logistic}, forecasting~\cite{zhang2004solving}, and decision-making~\cite{dawes1974linear} in many fields.

Moreover, linear regression forms the foundation for many advanced statistical and machine learning models~\cite{montgomery2021introduction}, including logistic regression~\cite{Cramer2002}, support vector machines~\cite{Cortes1995}, and generalized linear models~\cite{Nelder1972}.
Extensions of linear regression often aim to capture more complex relationships by introducing higher-order polynomial terms or additional nonlinear transformations. Modern developments in machine learning have enabled the training of deep and highly overparameterized models capable of modeling intricate patterns far beyond the reach of simple linear approaches. In particular, deep learning models can be interpreted as sophisticated forms of nonlinear regression~\cite{Zhou2024}, capable of approximating complex functions with high flexibility.

Despite its utility, linear regression struggles with modern high-dimensional datasets where only a small subset of variables is truly informative. For example, in medicine, identifying a few key Single Nucleotide Polymorphisms (SNPs) or blood biomarkers from large genomics and proteomics data is crucial for diagnosing specific diseases~\cite{Shastry2007, Olsson2016}. Similarly, in economics, the growing availability of large-scale administrative and private sector data presents both new opportunities and challenges for finding the most relevant factors~\cite{Einav2014}. Even in machine learning, deep networks can be pruned with little loss in accuracy~\cite{Han2015, Frankle2018}, and transformer models in natural language processing are trained to focus attention on just a few key tokens in a sentence for understanding context~\cite{Clark2019, Michel2019}.

Handling a large number of independent variables can make linear regression computationally intensive and prone to overfitting~\cite{babyak2004you, hawkins2004problem}. High correlations among predictors (multicollinearity) complicate the interpretation of regression weights and inflate standard errors, reducing reliability. Overfitting arises when models are overly complex and capture noise, while underfitting occurs when models are too simple to capture underlying patterns. Avoiding these pitfalls requires balancing model complexity through appropriate model selection and regularization~\cite{ref:BiasVariance}.

Feature selection and regularization techniques, such as Ridge~\cite{ref:Ridge} and LASSO regression~\cite{ref:LASSO}, help mitigate these challenges by constraining the solution space. Enforcing sparsity focuses the model on the most relevant variables, reduces the risk of overfitting in high-dimensional settings, and improves generalization. Additionally, sparse models tend to be more interpretable, offering clearer insights into variable importance~\cite{ng2011sparse, huben2024sparse}.

In this study, we revisit a fundamentally different approach: the garrote method, which explicitly incorporates variable selection through binary selection variables~\cite{ref:NNGarrote}. Specifically, we adopt its variational inference formulation—Variational Garrote (VG)~\cite{ref:VG}—and enhance it using modern automatic differentiation techniques. We systematically compare VG with the representative sparse regression methods, Ridge and LASSO, using a consistent set of evaluation metrics. Our results show that VG performs particularly well in highly sparse regimes and provides more consistent and robust variable selection across varying levels of sparsity.

This paper is organized as follows: In Section~\ref{sec:theory}, we present a unified introduction to sparse regression models and provide a detailed explanation of the VG. Section~\ref{sec:experiments} compares model performance using both fully controllable synthetic datasets and complex real-world data. Finally, Section~\ref{sec:conclusion} summarizes our findings and briefly discusses potential applications.

\section{Theory}
\label{sec:theory}
\subsection{Sparse regression}

Linear regression aims to determine the relationship between the input $x$ and the output $y$ based on $M$ paired samples of input and output data, $D=\{x^\mu, y^\mu\}_{\mu=1}^M$.
Specifically, we consider an $N$-dimensional input $x = (x_1, x_2, \cdots, x_N)$ and a scalar output $y$.
The objective is to find the weights $w_i$ that represent the linear relationship between the $i$-th explanatory variable $x_i$ and the output $y$ across the entire dataset.
This relationship is described by
\begin{equation}
    y^\mu = \sum_{i=1}^N w_i x_i^\mu + \xi^\mu,
\end{equation}
assuming that data pairs $(x^\mu, y^\mu)$ follow a centered distribution. The regression error $\xi^\mu$ is modeled as a Gaussian random variable with zero mean and variance $1/\beta$.

For overdetermined systems ($M > N$), the optimal set of weights that minimizes the regression error is given by ${w} = {x}^+{y}$, where ${}^+$ denotes the Penrose-Moore pseudoinverse~\cite{Penrose1955}, which is uniquely determined.
In contrast, for underdetermined systems ($M < N$), the solution is not unique.
To determine an optimal set of weights, additional constrains are required.
The underdetermined nature of the system causes the weights to overfit the training dataset.
This hence significantly increases the generalization error. 

Numerous approaches have been proposed to address overfitting by reducing the degrees of freedom of the weights. Most of these methods achieve this by imposing a regularization condition on the weights. By strengthening the regularization, less relevant input variables (or basis functions) are suppressed, effectively shrinking the corresponding regression weights.
Adjusting these constraints controls the number of relevant basis functions, helping to prevent overfitting in data-scarce, high-dimensional settings. It also offers insight into which basis functions are relevant, especially in underdetermined regression problems. An issue with some of these constraints is that they lead to weight shrinkage, even of relevant predictor variables, which can be an issue in severely underdetermined problems like genome-wide personalized risk score computations.

Optimal weights $w$ can be derived using maximum a posteriori (MAP) estimation based on Bayes' rule:
\begin{equation}
    P(w, \beta | D) = \frac{P(D|w, \beta)P(w)P(\beta)}{P(D)},
\end{equation}
assuming $w$ and $\beta$ are independent.
Here the likelihood for independent $\mu$ samples is defined as $P(D|w,\beta)=\prod_\mu P(x^\mu, y^\mu|w, \beta)$ with
\begin{equation}
    P(x^\mu, y^\mu | w, \beta) =  \sqrt{ \frac{\beta}{2\pi}} \exp \left[ -\frac{\beta}{2} \left(y^\mu - \sum_i  w_i x_i^\mu \right)^2 \right].
\end{equation}
The posterior probability of $w$ implies that the additional regularization, which constrains the distribution of the weights, corresponds to the prior distribution $P(w)$.
An unconstrained model with no inductive bias on the weights corresponds to a flat prior, $P(w) = {\rm const.}$, as in standard linear regression. 

A simple form of constraint is penalizing the $L_2$-norm of the weights, known as Ridge regression~\cite{ref:Ridge}. The corresponding prior distribution of the weights, $w = (w_1, w_2, \cdots, w_N)$, is a Gaussian distribution, $P(w) \propto \exp(-\lambda/2 \sum_i w_i^2)$.
The exact solution of this sparse regression is given by
${w} = \left({x^Tx +\lambda I}\right)^{-1} {x^Ty}$,
which can be interpreted as incorporating a Tikhonov regularizer in the matrix inversion. This regularizer does not shrink the weights toward zero; as a result, Ridge regression may reduce the generalization error but does not effectively distinguish irrelevant variables in the model.

Another constrained model is LASSO~\cite{ref:LASSO}, which penalizes the weights using the $L_1$-norm.
This regularization corresponds to a Laplace prior distribution of the weights, $P(w) \propto \exp(-\lambda \sum_i \vert w_i \vert)$, which has heavier tails compared to a Gaussian distribution.
Although LASSO does not have a closed-form analytical solution, it remains convex and can be efficiently solved using gradient descent or other optimization techniques~\cite{ref:LASSO, BeckTeboulle2009, Friedman2010, Boyd2011}. The key advantage of LASSO over Ridge regression is its ability to shrink irrelevant weights to exactly zero, resulting in sparse solutions and enabling the identification of relevant variables.
ElasticNet~\cite{ref:ElasticNet} combines the $L_1$ and $L_2$ regularizers, offering greater flexibility in controlling sparsity while benefiting from the strengths of both methods.
The level of sparsity can be adjusted by tuning the regularization strength $\lambda$ to either achieve a desired level of sparsity or minimize the validation error.

It is natural to extend the $L_p$-norm regularizer to cases where $p < 1$, as smaller values of $p$ can enhance sparsity. In particular, the $L_0$-norm, which represents the number of nonzero elements, is directly associated with controlling sparsity. However, $p = 1$ corresponds to the convex hull of the regularizer, making it more amenable to optimization. For $p < 1$, the regularizer becomes non-convex, and standard gradient descent methods do not guarantee a globally optimal solution.
Nonetheless, several methods have been proposed to impose model constraints analogous to the $L_0$-norm. A well-known approach in this class of regularizers is the non-negative garrote (NNG)~\cite{ref:NNGarrote}. NNG introduces a non-negative selection variable, $s = (s_1, s_2, \cdots, s_N)$ with $s_i \geq 0$, that satisfies the relationship:
\begin{equation}
\label{eq:VG}
  y^\mu = \sum_i s_i w_i x_i^\mu + \xi^\mu,  
\end{equation} 
where the weights $w_i$ are first estimated using ordinary linear regression. The selection variables $s_i$ are then constrained by $\sum_i s_i \leq t$.

Another method employs a Bayesian approach~\cite{ref:SpikeAndSlab, ref:BayesSpSl}, where the prior distribution $P(w)$ is modeled as a mixture of two Gaussian distributions--one narrow and one wide. The selection variable $s$ is then inferred as part of the model.
However, directly solving for the selection variable $s$ involves an exhaustive search, which becomes computationally infeasible for systems with a large degree of freedom $N$. Furthermore, computing the posterior distribution requires techniques such as Monte Carlo sampling or variational Bayesian approximation, adding to the computational complexity.

The VG~\cite{ref:VG} addresses these challenges by employing a Bayesian framework with a variational approximation to integrate out the selection variable. 
This method is analogous to finding an approximate energy landscape of a Sherrington-Kirkpatrick model using a variational approximation~\cite{mezard1987spin}. However, the VG requires careful scaling of the dataset and appropriate model initialization due to its lack of scaling properties.
The following section presents a detailed formulation of the VG model.

\subsection{Variational garrote}
The VG model considers binary selection variables $s_i = \{0, 1\}$ in Eq.~(\ref{eq:VG}),
where $s_i = 0$ indicates that the $i$-th explanatory variable $x_i$ does not contribute to explaining the output $y$, and $s_i = 1$ indicates that it does.

The posterior probability of the VG model is expressed as
\begin{equation}
    P(s, w, \beta|D) = \frac{P(D|s, w, \beta) P(s) P(w) P(\beta)}{P(D)}, 
\end{equation}
where the independence of $s$, $w$, and $\beta$ is assumed.
The likelihood of the dataset $(x^\mu, y^\mu)$ is given by
\begin{equation*}
    P(x^\mu, y^\mu |s, w, \beta) =  \sqrt{ \frac{\beta}{2\pi}} \exp \left[ -\frac{\beta}{2} \left(y^\mu - \sum_i  w_is_i x_i^\mu \right)^2 \right].
\end{equation*}
Furthermore, the VG assumes flat priors for $P(w)$ and $P(\beta)$.
However, it enforces sparsity in the selector variable $s$ with the following factorizable prior distribution:
\begin{equation}
    P(s) = \prod_{i} P(s_i) = \prod_i \frac{\exp(-\gamma s_i)}{1 + \exp(-\gamma)}.
\end{equation}
Therefore, increasing $\gamma$ leads to stronger sparsity.

The MAP estimate for the VG corresponds to minimizing the negative log-posterior.
In particular, the VG considers a posterior that marginalizes over the selection variable $s$.
To achieve this, it employs the following variational bound:
\begin{align}
- \ln \sum_s P(s, w, \beta | D) \leq - \sum_s Q(s) \ln \frac{P(s, w, \beta | D)}{Q(s)} \equiv F,
\end{align}
which is derived using Jensen's inequality.
The variational method approximates probability equations by calculating a bound using a proxy probability distribution $Q(s)$, which is employed when direct calculations with the original distribution $P(s)$ are computationally challenging.
Here, a simple Bernoulli distribution is adopted as the proxy distribution:
\begin{equation}
    Q(s) = \prod_i Q_i(s_i) = \prod_i m_i s_i + (1-m_i) (1-s_i),
\end{equation}
where $m_i$ controls the probabilities for $s_i = 0$ or $s_i = 1$.

Then, the variational bound, referred to as the free energy, is expressed as
\begin{align}
F(m, w, \beta) = & \beta E(m, w) - S(m) \nonumber \\
&- \gamma \sum_i m_i  - \frac{M}{2}\ln\frac{\beta}{2\pi} + C(\gamma).
\label{eq:VGLoss}
\end{align}
The first data-driven loss, $E(m, w)$, is an energy-like term representing reconstruction errors:
\begin{align}
E(m, w) \equiv & \frac{1}{2}\sum_\mu \left( y^\mu-\sum_i m_i w_i x^\mu_i \right)^2 \nonumber \\
& + \frac{1}{2} \sum_\mu \sum_i m_i(1-m_i)\left( w_i{x_i^\mu} \right)^2,
\label{eq:VGLoss2}
\end{align}
where the second term arises from the property $s_i^2 = s_i$ for the binary selection variable.
The second loss, $S(m)$ can be interpreted as an entropy-like term:
\begin{equation}
    S(m) \equiv \sum_i h(m_i),
\end{equation}
where $h(m_i) = - m_i \ln m_i - (1-m_i) \ln (1-m_i)$ is the Shannon entropy function. Here, the entropy term encourages an equal distribution, $m_i = 1/2$, for $s_i = 0$ and $s_i = 1$.
The temperature-like parameter $\beta$ acts as a balancing factor between the energy $E(m,w)$ and the entropy $S(m)$.
In the low temperature limit (large $\beta$), the VG prioritizes minimizing $E(m, w)$, while in the high temperature limit (small $\beta$), it focuses on maximizing $S(m)$.
The additional terms serve as regularization for $m$ and $\beta$. 
The last term, $C(\gamma)$, is a constant independent of $m$, $w$, and $\beta$.

The original VG model uses the minimization conditions $\partial F/\partial m_i=0$, $\partial F/\partial w_i=0$, and $\partial F/\partial \beta=0$ to iteratively update and obtain optimal solutions for $(m, w, \beta)$~\cite{ref:VG}.
Here, we observe that the third equation,
\begin{equation}
    \frac{\partial F}{\partial \beta} = E(m, w) - \frac{M}{2\beta} = 0,
    \label{eq:optim}
\end{equation}
exhibits very slow convergence when the solution is far from the optimum.
By using the equality $\beta = M/2E(m,w)$ from this equation, $\beta$ can be eliminated in Eq.~(\ref{eq:VGLoss}), resulting in the following expression:
\begin{widetext}
\begin{align}
F(m, w) =& \frac{M}{2} \ln \left[ \sum_\mu ( y^\mu-\sum_i m_iw_i x^\mu_i )^2+ \sum_\mu \sum_i m_i(1-m_i) (w_i{x_i^\mu})^2 \right] \nonumber \\
&  + \sum_i m_i \ln m_i + (1-m_i) \ln (1-m_i) - \gamma \sum_i m_i,
\label{eq:VGLoss2}
\end{align}   
\end{widetext}
where constant terms such as $M/2-M/2\ln (M/2\pi) + C(\gamma)$ are omitted, as they are irrelevant for the optimization of $m$ and $w$. 
This approach of eliminating $\beta$ significantly accelerates convergence.
In the analytic form of $F(m, w)$, $m_i$ can be interpreted as mean-field-like values of the selection variable $s_i$.
Henceforth, since $m_i=0$, similar to $s_i=0$, effectively disables the $i$-th explanatory variable $x_i$, we refer to $m_i$ as a {\it mask} variable.

The minimization of $F(m, w)$ with respect to $m$ and $w$ can be efficiently performed using advanced gradient descent methods available in modern machine learning frameworks.
Masks are initially set to $m_i = 1$, and weights are sampled from a normal distribution with zero mean and unit standard deviation. 
The loss is then calculated using Eq.~(\ref{eq:VGLoss2}) and optimized with the AdamW optimizer, employing an adaptive learning rate schedule starting from 0.03. Training is iterated until the learning rate decreases below $10^{-6}$, ensuring convergence.

\section{Experiments}
\label{sec:experiments}
We compare three representative sparse regression models—Ridge, LASSO, and VG—under consistent settings and evaluation metrics. Their performance is assessed on both synthetic and real-world datasets.

\subsection{Experiments on synthetic datasets}
\subsubsection{Experimental setup}
\begin{figure}[t]
\includegraphics[width=\columnwidth]{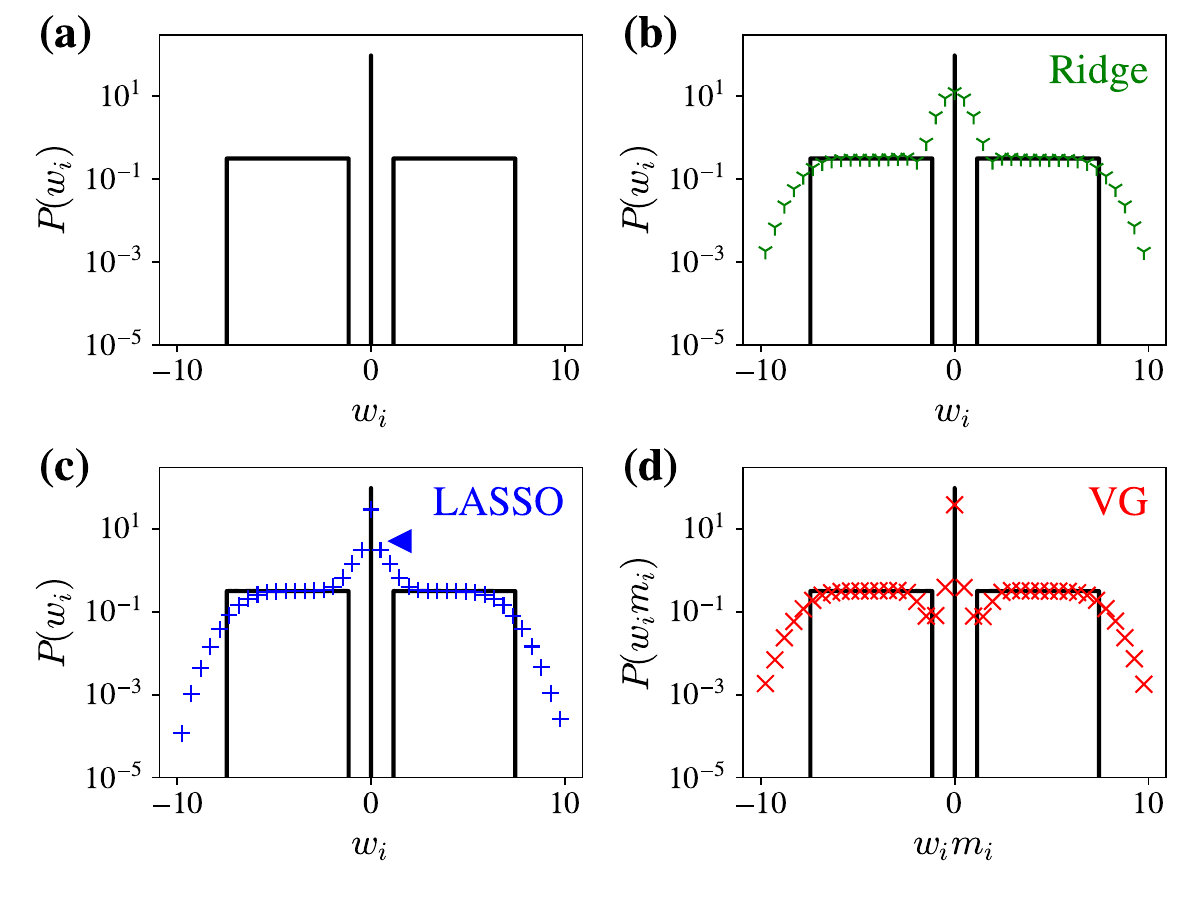}
\centering
    \caption{Inference of spike-and-slab distributions. (a) A spike-and-slab distribution characterized by a sharp peak at $w_i = 0$ and two uniform components over certain positive and negative ranges of $w_i$. Given the true distribution, the optimally inferred distributions from three sparse regression models are shown as symbols: (b) Ridge (green ``{\tiny Y}"); (c) LASSO (blue ``{\tiny +}"); and (d) Variational Garrote (VG, red ``{\tiny X}"). In the LASSO plot, the separation between the sharp peak and the broader distribution--referred to as the ``elbow"--is marked by a triangle. The VG model jointly infers both the weight $w_i$ and a mask variable $m_i$.
    }
\label{fig:Sparsity}
\end{figure}

We begin by examining sparse regression using a synthetic dataset that allows for systematic control. This dataset is designed to emulate real-world conditions, where only a small subset of variables has a significant impact on the target.
Since regression performance can depend on the true distribution of regression weights, we adopt a spike-and-slab distribution that provides flexible control of variable sparsity via a single density parameter, $\rho_{\text{data}}$. The distribution is defined as:
\begin{equation}
    P(w_i) = \begin{cases}
1-\rho_{\text{data}} &\text{$w_i = 0$}\\
\frac{1}{2} \rho_{\text{data}} &\text{$1<|w_i|< \bar{w}$} \\
0 &\text{otherwise},
\end{cases}
\end{equation}
where the upper bound $\bar{w} = \sqrt{12/\rho_{\text{data}} - 3/4} - 1/2$ is chosen to ensure that the variance of $w_i$ is finite and normalized (here, set to 2). 
The resulting shape of the spike-and-slab distribution $P(w_i)$ is shown in Fig.~\ref{fig:Sparsity}(a).
The parameter $\rho_{\text{data}}$ governs the balance between the spike (zero weights) and the slab (nonzero weights), and thus directly controls the variable sparsity level of the dataset.
For instance, $\rho_{\text{data}} = 0.5$ implies that half of the variables are relevant (nonzero weights), while the other half are irrelevant (zero weights).

Using weights $w_i$, drawn from $P(w_i)$, we generate a regression dataset. The target output is calculated according to $\hat{y}^\mu = \sum_i w_i x_i^\mu + \xi^\mu$, where each input variable $x_i^\mu$ is independently sampled from a normal distribution with zero mean and unit standard deviation. While this setup allows for controlled analysis, it differs from real-world scenarios in which the input vector $x$ often exhibits complex correlations and structure.
To simulate realistic measurement conditions, we introduce additive noise $\xi^\mu$, with its amplitude calibrated to achieve a signal-to-noise ratio of $\langle (\sum_i w_i x_i^\mu)^2 \rangle / \langle (\xi^\mu)^2 \rangle = 3$, where $\langle \cdot \rangle$ denotes the ensemble average. This ensures the noise contributes a non-negligible but controlled level of uncertainty, preserving the interpretability of the regression task.
The synthetic dataset $\{x^\mu, \hat{y}^\mu\}$ is now prepared. The true regression weights $w_i$ are hidden, and the objective is to infer $w_i$ using only the observed dataset $\{x^\mu, \hat{y}^\mu\}$.
For clarity, we refer the true prior distributions of weights as ``teacher'' distributions, and the distributions estimated by the regression models as ``student'' distributions.
In this task, we set the dimensionality of $x$ to $N=256$ and the number of samples to $M=256$, achieving a balanced data-to-dimension ratio of $M/N = 1$. 
This ratio was deliberately chosen to evaluate the performance and robustness of the regression models under a critical regime where the number of features matches the number of observations.

\subsubsection{Variable sparsity}

To enable a fair comparison of sparse models, we face a challenge: the regularization terms differ in form, each with its own sparsity-controlling parameters. To address this, we introduce a unified measure for variable sparsity that can be applied consistently across all three models of Ridge, LASSO, and VG.

In the VG model, sparsity can be computed explicitly, as the mask variable $m_i$ is inherently integrated into the model. It serves as the mean-field value of the binary selection variable $s_i$.
The level of variable sparsity is determined by the density of active elements (i.e., those with $m_i \approx 1)$ among the total of $N$ variables, and is quantified as:
\begin{equation}
    \rho_{\text{model}} = \frac{1}{N} \sum_{i=1}^N m_i.
\label{eq:rho}
\end{equation}

However, the Ridge and LASSO models do not explicitly include mask or selection variables; therefore, the mask $m_i$ must be inferred from the corresponding weights $w_i$, based on their relative magnitudes.
To illustrate this more clearly, we present a visual example in Fig.~\ref{fig:Sparsity}. Using a teacher weight distribution modeled by a spike-and-slab prior with $\rho_{\text{data}}=47/256$, we plot the best student weight distributions for the three regression models, each using its optimal regularization strength.

For LASSO, the student weight distribution exhibits a bimodal structure: irrelevant variables cluster tightly around zero, forming a sharp peak, while relevant variables are more broadly dispersed. Ideally, irrelevant weights would shrink exactly to zero~\cite{ee685348-c1b0-32a8-849d-4be4de07f8ad}; however, due to finite numerical precision and the presence of noise, they tend to remain close to zero rather than reaching it exactly.
To distinguish relevant from irrelevant variables, we assign $m_i =1$ (and $m_i=0$ otherwise) to those weights whose magnitudes exceed the “elbow” of the distribution. This threshold is determined by fitting the weight histogram to a zero-centered binary mixture model~\cite{Zhou2010ThresholdedLF}. In this example, the elbow point is indicated by a blue triangle (Fig.~\ref{fig:Sparsity}(c)).

In Ridge regression, all mask values are theoretically $m_i=1$, since the weights never shrink to zero and remain finite. 
To identify relevant variables, we introduce a lower bound $\underline{w}$; any weight $w_i$ with $|w_i| \geq \underline{w}$ is considered indicative of a relevant feature $x_i$ and assigned $m_i=1$; otherwise $m_i=0$.
To determine $\underline{w}$, we first fit a standard (unregularized) linear regression and collect its smallest absolute weights across the ensemble. The average of these minimal weights serves as a representative uncertainty level, which we adopt as our lower bound.

Once the mask variables $m_i$ are defined for both LASSO and Ridge regression, the corresponding variable sparsity levels can be computed using Eq.~(\ref{eq:rho}).

\begin{figure*}[t]
\includegraphics[width=0.9\columnwidth]{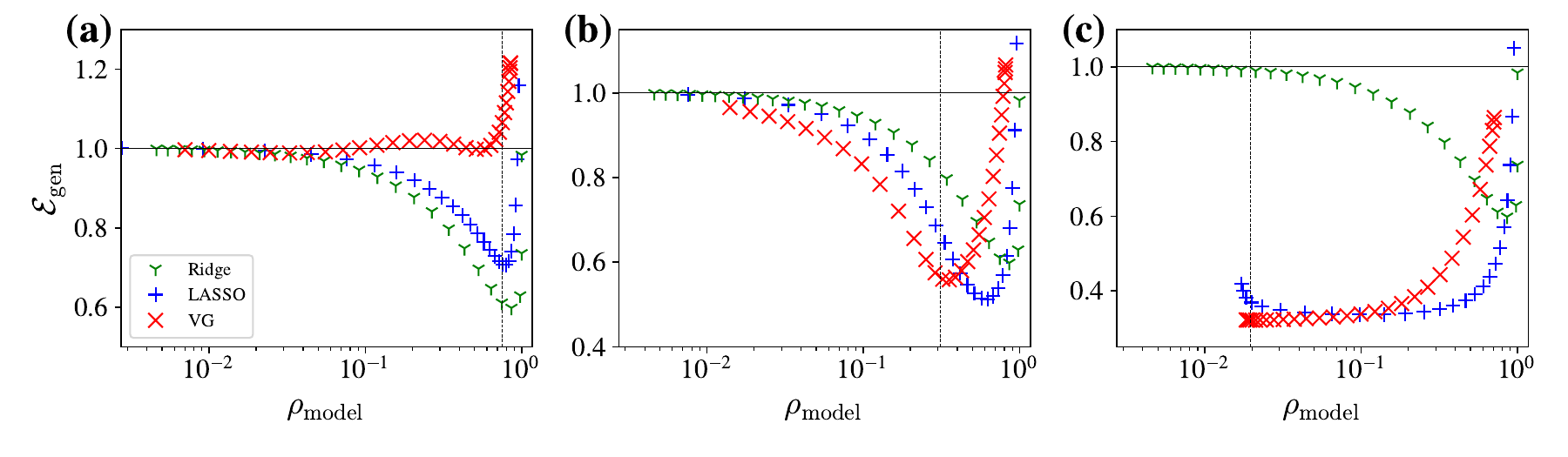}
\centering
    \caption{Generalization error of sparse regression models. Errors for Ridge (green ``{\tiny Y}"), LASSO (blue ``{\tiny +}"), and Variational Garrote (VG, red ``{\tiny X}") are shown as a function of sparsity level, tuned via the model density $\rho_{\text{model}}$ of relevant variables. Results are given for true sparsity levels: (a) $\rho_{\text{data}} = 192/256$, (b) $80/256$, and (c) $5/256$. Dotted vertical lines indicate the corresponding $\rho_{\text{data}}$.
    }
\label{fig:SynthRegression}
\end{figure*}

\subsubsection{Generalization performance}
We now evaluate the performance of sparse regression models under varying numbers of relevant variables.
The primary performance metric is the generalization error, which quantifies the model's ability to predict $y$ for previously unseen input data $x$. 
It is defined as the normalized distance between the estimated values $y^\mu$ and true values $\hat{y}^\mu$:
\begin{equation}
    \mathcal{E}_{\text{gen}} = \left \langle \sqrt{\frac{\sum_\mu \left( y^\mu - \hat{y}^\mu \right)^2}{\sum_\mu \left( \hat{y}^\mu \right)^2}}  \right \rangle,
\end{equation} %% To be reviewed after
where the ensemble average $\langle \cdot \rangle$ is taken over multiple test datasets $\{x^\mu, \hat{y}^\mu\}$.
This metric evaluates model performance on unseen data, not the training set, and thus serves as a measure of generalizability.

We analyze three regimes of teacher weight distributions corresponding to different levels of true variable density $\rho_{\text{data}}$.
In a high density regime ($\rho_{\text{data}} = 192/256$), Ridge regression achieves the lowest generalization error $\mathcal{E}_{\text{gen}}$ when the model’s inferred density level $\rho_{\text{model}}$ closely matches the ground-truth density $\rho_{\text{data}}$ (Fig.~\ref{fig:SynthRegression}(a)).
LASSO also performs well near $\rho_{\text{model}} \approx \rho_{\text{data}}$, though its performance is slightly inferior to Ridge in this regime.
This alignment supports the validity of our definition of $\rho_{\text{model}}$ as a reasonable measure of variable sparsity.
Notably, $\mathcal{E}_{\text{gen}}$ increases when the model includes either too few or too many variables. However, overestimating the number of relevant variables—i.e., using more than necessary—results in a more substantial degradation in performance than underestimating them.
In this high density regime, VG performs poorly, particularly when it overestimates the number of relevant variables.

Next, in the intermediate density regime ($\rho_{\text{data}} = 80/256$), LASSO achieves the lowest $\mathcal{E}_{\text{gen}}$ (Fig.~\ref{fig:SynthRegression}(b)).
However, both VG and Ridge also demonstrate competitive performance.
In this regime, the optimal $\rho_{\text{model}}$ for LASSO and Ridge deviates slightly from the true $\rho_{\text{data}}$. 
In contrast, VG--which explicitly models the mask variables--still achieves optimal performance when $\rho_{\text{model}} \approx \rho_{\text{data}}$.

Finally, in the very low density regime ($\rho_{\text{data}} = 5/256$), VG achieves the lowest $\mathcal{E}_{\text{gen}}$, although LASSO also performs competitively (Fig.~\ref{fig:SynthRegression}(c)).
VG reaches optimal performance near $\rho_{\text{model}} \approx \rho_{\text{data}}$, but its performance remains stable across a broad range of sparsity levels deviating from the ground truth—a behavior similarly observed for LASSO.
In contrast, Ridge performs poorly in this highly sparse setting, indicating its limitations when the number of relevant variables is very small.

In summary, as the number of relevant variables in the data decreases, sparser models—progressing from Ridge to LASSO to VG—achieve increasingly better regression performance.

\begin{figure}[t]
\includegraphics[width=1.\columnwidth]{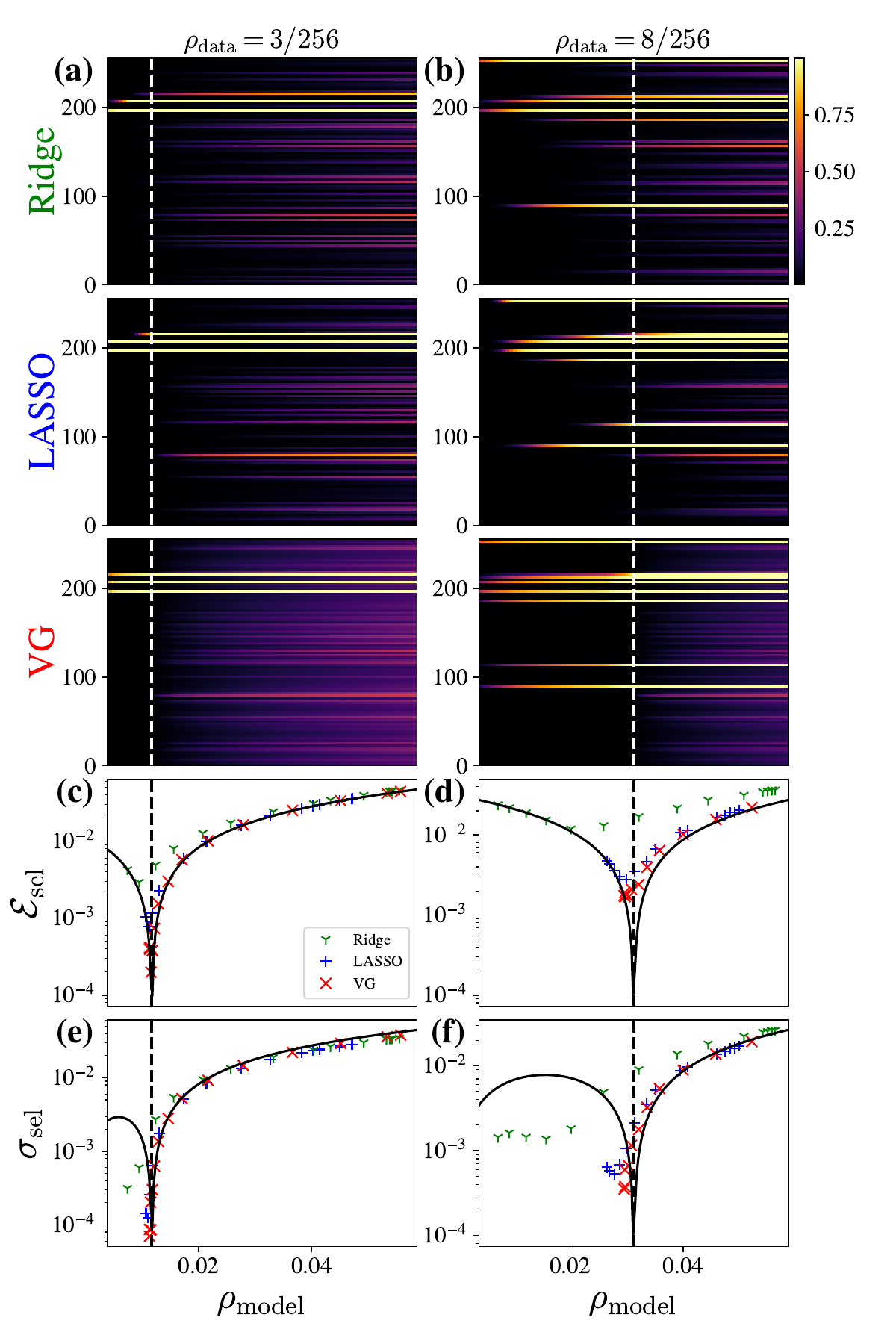}
\centering
    \caption{Variable selection in sparse regression models. Ensemble averages $\langle m_i \rangle$ of the $i$-th mask among 256 variables are plotted as a function of the model sparsity degree $\rho_{\text{model}}$ for true densities (a) $\rho_{\text{data}} = 3/256$ and $\rho_{\text{data}} = 8/256$. Selected variable appear brighter, while unselected variable appear darker. The corresponding selection errors for Ridge (green ``{\tiny Y}"), LASSO (blue ``{\tiny +}"), and Variational Garrote (VG, red ``{\tiny X}") are shown in (c) and (d), and the selection uncertainties in (e) and (f). Vertical dotted lines mark $\rho_{\text{data}}$. Results are averaged over 20,000 ensembles. Solid black lines in (c-f) represent mean-field theoretical estimates.}
\label{fig:SynthRegression2}
\end{figure}

\subsubsection{Variable selection performance}
We next examine how each sparse model's variable selection evolves as the permitted density $\rho_{\text{model}}$ increases.
By focusing on low-density regimes, we can assess each model's capability to accurately identify the true relevant variables.
Figures~\ref{fig:SynthRegression2}(a) and (b) visually illustrate how the three sparse regression models progressively select relevant features.
The ensemble-averaged mask values $\langle m_i \rangle$ indicate which variables or features are selected from a total of 256 features as $\rho_{\text{model}}$ increases.
In the highly sparse case ($\rho_{\text{data}} = 3/256$), when $\rho_{\text{model}}$ is too low, Ridge or LASSO typically identify only one or two of the three true relevant variables, often missing at least one.
This indicates a tendency for both models to prioritize certain features, resulting in biased selection.
This behavior can be attributed to the convex nature of their loss functions, which leads to uniquely ordered solutions for the weights $w_i$.
As sparsity increases, this ordering encourages the selection of a single dominant variable, even when multiple equally relevant alternatives exist.
In contrast, VG reliably identifies all three relevant variables by assigning soft mask values with $0< m_i < 1$, enabling more flexible selection.
This key property underlies the lower generalization error observed for VG models in the oversparsified regime, compared to Ridge and LASSO (Fig.~\ref{fig:SynthRegression}(c)).
We also observe that all three models undergo a sharp transition in the number of selected relevant variables once $\rho_{\text{model}}$ exceeds approximately $\rho_{\text{data}}$.

To examine this behavior more quantitatively, we assess each model's selection performance by comparing the true selection variables ${s}_i$ with the model's estimated mask values ${m}_i$.
To quantify the mismatch between the true and estimated selections, we define the selection error as: 
\begin{equation}
\label{eq:e_sel}
    {\mathcal E}_{\text{sel}} = \frac{1}{N} \sum_{i=1}^N  \Big \langle {s}_i (1-{m}_i) + (1-{s}_i) {m}_i \Big \rangle,
\end{equation}
which measures the average rate of incorrect selections (false negatives and false positives) across the ensemble.
In this setting, VG achieves the lowest selection error ${\mathcal E}_{\text{sel}}$, followed by LASSO, with Ridge performing the worst.
For all three models, ${\mathcal E}_{\text{sel}}$ reaches minimum when the inferred density $\rho_{\text{model}}$ closely aligns with the true density $\rho_{\text{data}}$ (see Fig.~\ref{fig:SynthRegression2}(c) for $\rho_{\text{data}} = 3/256$ and Fig.~\ref{fig:SynthRegression2}(d) for $\rho_{\text{data}} = 8/256$).
Note that in practice it is difficult to reduce $\rho_{\text{model}}$ below $\rho_{\text{data}}$ by tuning the regularization parameter in the LASSO and VG models. Consequently, the visualization of variable selection in Figs.~\ref{fig:SynthRegression2}(a) and (b) within these highly sparse regions is based on extrapolation.

Under a simple mean-field assumption with variable independence, this minimum behavior of ${\mathcal E}_{\text{sel}}$ can be understood intuitively.
In the case of under-selection ($\rho_{\text{model}} < \rho_{\text{data}}$), the sparse model identifies $N\rho_{\text{model}}$ relevant variables, which is fewer than true $N\rho_{\text{data}}$ relevant variables. This leads to a per-variable selection error of 
${\mathcal E}_{\text{sel}} = \rho_{\text{data}} - \rho_{\text{model}}$.
Conversely, in the case of over-selection ($\rho_{\text{model}} > \rho_{\text{data}}$), the error becomes ${\mathcal E}_{\text{sel}} = \rho_{\text{model}} - \rho_{\text{data}}$.
We also provide a more rigorous derivation for this behavior in Appendix~\ref{sec:appendix2}.
The mean-field estimate is depicted by the solid line in Figs.~\ref{fig:SynthRegression2}(c) and (d).

Next, we examine the variability in variable selection across different realizations of the synthetic dataset.
This variability is quantified by the ensemble-averaged uncertainty of the selection mask values over multiple test datasets:
\begin{equation}
    \mathcal{\sigma}_{\text{sel}} = \frac{1}{N} \sum_{i=1}^N \big \langle  m_i \big \rangle \big \langle 1 - m_i \big \rangle,
    \label{eq:uncertainty}
\end{equation}
which reflects the consistency of variable selection within the ensemble.
Analogous to the binomial distribution, the uncertainty of each averaged mask value is maximized at $\langle m_i \rangle = 0.5$, and minimized at $\langle m_i \rangle = 0$ or $\langle m_i \rangle = 1$.
The overall selection uncertainty is smallest when $\rho_{\text{model}} \approx \rho_{\text{data}}$, as shown in Figs.~\ref{fig:SynthRegression2}(e) and (f).
Intuitively, uncertainty increases when the model selects either substantially fewer or substantially more relevant variables than the true number.
This property can be particularly useful for identifying the appropriate number of relevant variables in real-world problems where the true variable set is unknown.

Using the mean-field approximation, we can further derive a theoretical estimate for the selection uncertainty:
\begin{align}
    \sigma_{\text{sel}}&=
    \begin{cases}
        \dfrac{\rho_{\text{model}}}{\rho_{\text{data}}} \big(\rho_{\text{data}} - \rho_{\text{model}} \big), & \text{for $\rho_{\text{model}}<\rho_{\text{data}}$},\\[2mm]
        \big(\rho_{\text{model}} - \rho_{\text{data}}\big) \dfrac{1-\rho_{\text{model}}}{1-\rho_{\text{data}}}, & \text{for $\rho_{\text{model}}>\rho_{\text{data}}$}.
    \end{cases}    
    \label{eq:kernel}
\end{align}
The theoretical estimate is shown as solid lines in Figs.~\ref{fig:SynthRegression2}(e) and (f), with the detailed derivation provided in Appendix~\ref{sec:appendix1}.

\begin{figure}[t]
\includegraphics[width=1.\columnwidth]{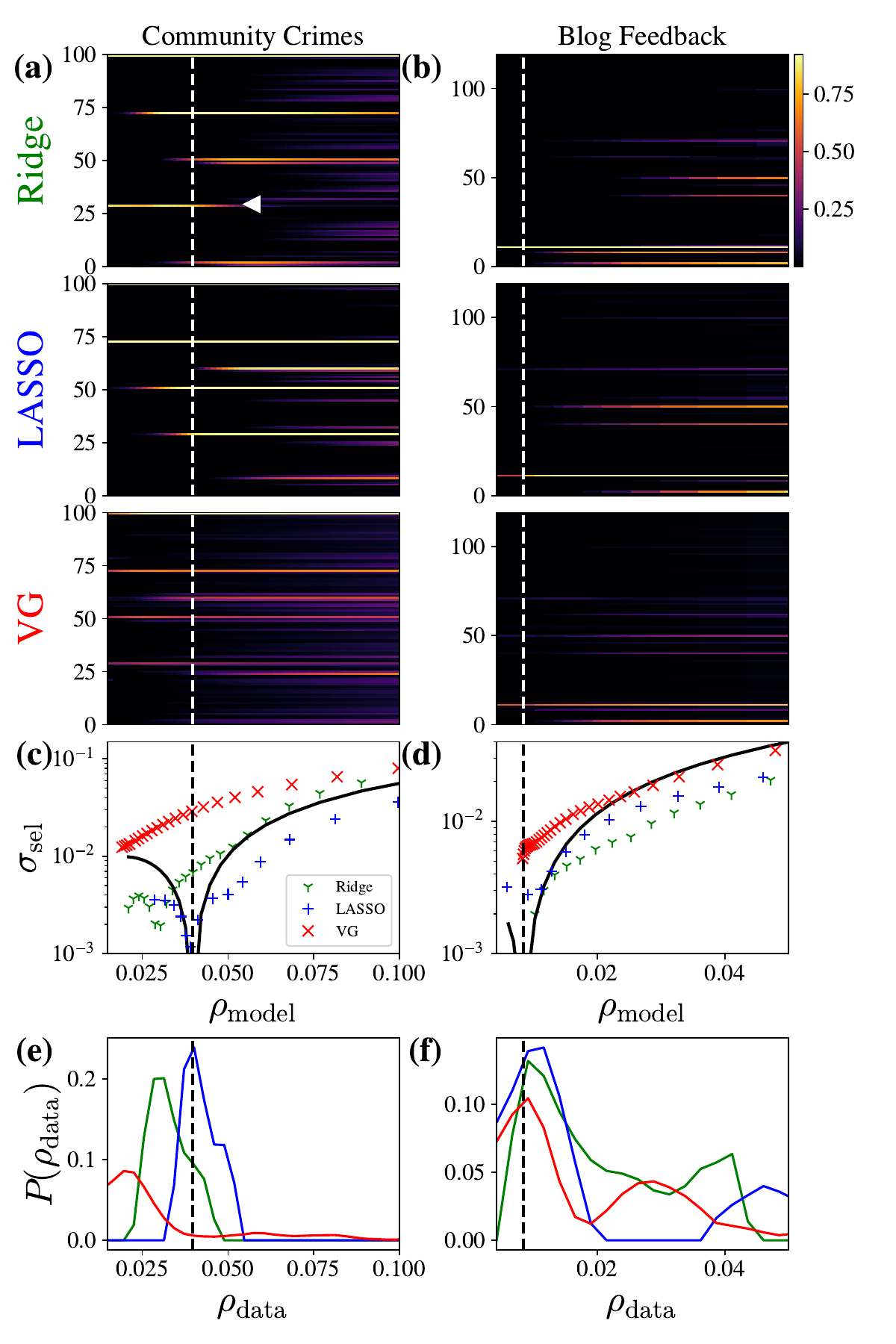}
\centering
    \caption{
        Variable selection in real-world datasets. (a-b) Ensemble averages $\langle m_i \rangle$ of the $i$-th selection mask across 101 variables in the Community Crimes (CC, left column) dataset and 120 variables in the Blog Feedback (BF, right column) dataset, respectively. Brighter values indicate selected variables, while darker values indicate unselected variables. (c-d) Selection uncertainties for Ridge (green “{\tiny Y}”), LASSO (blue “{\tiny +}”), and Variational Garrote (VG, red “{\tiny X}”), averaged over 20,000 ensembles. Solid black lines denote mean-field theoretical estimates. (e-f) Inferred probability distributions of the true data sparsity $\rho_{\text{data}}$ for CC and BF, respectively. Vertical dotted lines indicate the point estimate of $\rho_{\text{data}}$ from LASSO.
    }
\label{fig:RWSelection}
\end{figure}

\subsection{Experiments on real-world datasets}
We evaluate our findings on synthetic datasets to validate the model’s robustness and practical applicability to real-world scenarios. These real-world datasets are often characterized by noise, missing values, strong implicit correlations, and complex nonlinear relationships, reflecting common challenges in practical applications. By examining such conditions, we aim to assess the model’s ability to capture underlying structures and effectively manage real-world complexities.

We chose two datasets from the UCI Machine Learning Repository. The first is the Communities and Crimes dataset~\cite{ref:CCDataset} (CC), a multivariate regression dataset that combines socioeconomic data, law enforcement data, and crime data for communities in the United States. This dataset contains over 100 continuous variables with moderate correlations, making it suitable for evaluating the model's ability to handle collinearity between variables. 
The dataset consists of $M=2215$ instances (communities), 125 attributes (socioeconomic and law enforcement features), and 18 target variables (crimes counts by type). We preprocessed the data for regression by filtering out attributes with missing or erroneous values, retaining $N=101$ of the original 125 attributes. From the 18 target variables, we selected the most significant one based on the highest total count. As the feature data is positive and follows a heavy-tailed distribution, we applied a logarithmic transformation, standardized each value, and set a lower cutoff of $-3$ to address issues with zero values.

The second dataset is the Blog Feedback dataset~\cite{ref:BlogFeedback} (BF), also from the UCI Machine Learning Repository. This multivariate regression dataset predicts a blog's future comment activity based on various features, including statistics (e.g., length of a post) and qualitative attributes (e.g., the presence of specific words), totaling 280 features. The target variable is the number of feedback instances on a blog post. 
The BF dataset comprises $M=60021$ blog instances with continuous variables that exhibit non-global collinearity patterns, similar to those in the CC dataset.
We selected 60 quantitative variables, excluding categorical ones, and augmented the dataset by applying a signed logarithm transformation to each variable, resulting in $N=120$ features.

To evaluate performance, we created overlapping cross-validation datasets by randomly sampling $r M$ data points for training and using the remaining $(1-r) M$ data as the test set, with $r = 0.15$ for the CC dataset and $r = 0.01$ for the BF dataset.
A smaller $r$ was chosen for the BF dataset to maintain a comparable training set size, as it contains far more total instances than the CC dataset.

The regression results for each dataset are shown in Fig.~\ref{fig:RWSelection}.
Similar to the synthetic dataset, Ridge and LASSO tend to select one or two dominant variables when $\rho_{\text{model}}$ is very low in the CC dataset (Fig.~\ref{fig:RWSelection}(a)).
In contrast, VG selects relevant variables consistently and robustly across different $\rho_{\text{model}}$ levels. 
However, this consistency and robustness observed in VG are less pronounced for the BF dataset (Fig.~\ref{fig:RWSelection}(b)).
As $\rho_{\text{model}}$ increases, Ridge occasionally exhibits inconsistent behavior: a variable selected at a lower $\rho_{\text{model}}$ is later excluded at a higher $\rho_{\text{model}}$ (see the triangle in Fig.~\ref{fig:RWSelection}(a)).
This suggests that features initially deemed highly relevant are sometimes discarded when the model is allowed to include more variables, indicating instability in the selection process.
VG, by contrast, does not exhibit this behavior, demonstrating more consistent and robust variable selection across varying levels of $\rho_{\text{model}}$.

For real-world datasets, unfortunately, ${\mathcal E}_{\text{sel}}$ cannot be explicitly evaluated because the true selection variable $s_i$ is intractable. Nevertheless, the uncertainty ${\sigma}_{\text{sel}}$ of the selection mask values can still be estimated (Figs.~\ref{fig:RWSelection}(c) and (d)). By comparing the estimated ${\sigma}_{\text{sel}}$ with the expression in Eq.~(\ref{eq:kernel}), we can inversely infer $\rho_{\text{data}}$ (Figs.~\ref{fig:RWSelection}(e) and (f)). The detailed calculation of $P(\rho_{\text{data}})$ is provided in the Appendix~\ref{sec:appendix2}.
Based on this inference of data sparsity, the estimated number of relevant variables $N\rho_{\text{data}}$ is 3 for Ridge, 4 for LASSO, and 2.5 for VG in the CC dataset, while it is 1 across all models in the BF dataset.
In the CC dataset, the top three community-level variables consistently identified across models for predicting crime occurrence are: (i) {\it population}, (ii) {\it number of households under poverty}, and (iii) {\it number of kids born never married}. These features are primarily extrinsic factors reflecting population demographics. 
In the BF dataset, the single most relevant variable for predicting blog feedback events is {\it log maximum number of traceback links within 24 hours to 48 hours}.
Overall, these findings highlight the utility of selection uncertainty in identifying relevant variables across different datasets.

\section{Conclusions}
\label{sec:conclusion}
Sparse regression is a key technique for identifying relevant variables among many inputs, particularly in the era of big data. In this study, we evaluate representative sparse regression methods—Ridge, LASSO, and Garrote—using a consistent set of evaluation metrics. Special attention is given to the Variational Garrote (VG)~\cite{ref:VG}, a powerful yet underutilized method in both the physics and machine learning communities.

We begin with a synthetic dataset, where the nature and size of the data are fully controlled. Our results show that Ridge, LASSO, and VG all achieve accurate generalization and variable selection when the variable sparsity of the model aligns with that of the underlying data.
However, in sparser regimes with fewer relevant variables, VG outperforms LASSO, while Ridge performs the worst.
Furthermore, when the level of variable sparsity is systematically varied, VG consistently and robustly selects the relevant variables.
In contrast, Ridge often excludes variables that were selected under sparse conditions once more variables are permitted under denser settings.
When more variables than necessary are included, the uncertainty in variable selection increases sharply---particularly for VG, which exhibits a clear transition. 
Therefore, we suggest that this transition point can serve as an effective indicator for estimating the true number of relevant variables.

We then extend our analysis to real-world datasets, guided by insights from the synthetic data. VG continues to perform well, achieving the low generalization error in highly sparse regimes. It also exhibits a sharp transition in selection uncertainty, effectively identifying key relevant variables among many inputs in real-world settings.

However, some differences emerge. Ridge shows more pronounced inconsistency in variable selection compared to the synthetic case. Additionally, unlike in the synthetic dataset—where Ridge typically performs worst in sparser regimes—it achieves low generalization error even in sparser settings. This shift may stem from the inherent correlations and complex structure of input variables in real-world data.

In conclusion, VG is a compelling model for sparse regression, particularly in very sparse regimes, as it consistently and robustly identifies a small number of key relevant variables among many. Given the central role of sparse modeling in numerous domains, this approach has a wide range of applications~\cite{HansenStahlhutHansen2013, AndersenHansenHansen2013}.

For instance, in compressed sensing, our method aligns closely with the goal of recovering sparse signals from noisy data. Applications include denoising sparse signals or reconstructing full images from sinograms, a common challenge in the context of the Radon transform~\cite{Donoho2006, Lustig2007}.

In deep learning, weight distributions often shift from heavy-tailed to dense depending on layer depth or activation patterns~\cite{Martin2021, MartinMahoney2020, MengYao2023}. Incorporating layer-specific priors and regularization strategies may help improve model performance. Furthermore, the selective capabilities of LASSO and VG point to the potential for layerwise pruning, offering a way to enhance both the efficiency and interpretability of neural networks~\cite{Louizos2017, Louizos2018}.

\section*{Acknowledgments}
This research was supported (V.P.) by the Intramural Research Program of the National Institute of Diabetes and Digestive and Kidney Diseases (NIDDK) within the National Institutes of Health (NIH). The contributions of the NIH author(s) are considered Works of the United States Government. The findings and conclusions presented in this paper are those of the author(s) and do not necessarily reflect the views of the NIH or the U.S. Department of Health and Human Services. This research was supported by the Creative-Pioneering Researchers Program through Seoul National University, and the National Research Foundation of Korea (NRF) grant (Grant No. 2022R1A2C1006871) (J.J.).

\appendix
\section{Mean-field derivation}
\label{sec:appendix1}

We present a more rigorous derivation.
In the under-selection case, we set $m_i = 0$ for all irrelevant variables ($s_i = 0$);
in other words, we assume the absence of false positives.
Then, Eq.~(\ref{eq:e_sel}) becomes
\begin{align}
    {\mathcal E}_{\text{sel}} &= \frac{1}{N} \sum_{i=1}^N  {s}_i \Big(1- \langle {m}_i \rangle \Big) \nonumber \\
    &= \frac{1}{N} \sum_{i=1}^N  {s}_i \Big(1- \frac{\sum_i m_i}{\sum_i s_i} \Big) \nonumber \\
    &= \frac{1}{N} \sum_{i=1}^N  {s}_i - \frac{1}{N} \sum_{i=1}^N  {m}_i \nonumber \\
    &= \rho_{\text{data}} - \rho_{\text{model}}.
\end{align}
In the second line, we apply the mean-field assumption that each mask value $\langle m_i \rangle$ for a relevant variable equals the ratio between the total number of selected relevant variables, $\sum_i m_i$, and the total number of true relevant variables, $\sum_i s_i$. 

Conversely, in the over-selection case, we set $m_i = 1$ for all relevant variables ($s_i = 1$); in other words, we assume the absence of false negatives.
Then, Eq.~(\ref{eq:e_sel}) becomes
\begin{align}
    {\mathcal E}_{\text{sel}} &= \frac{1}{N} \sum_{i=1}^N  (1 - {s}_i) \langle {m}_i \rangle \nonumber \\
    &= \frac{N - \sum_i s_i }{N} \frac{\sum_i m_i - \sum_i s_i}{N - \sum_i s_i} \nonumber \\
    &= \rho_{\text{model}} - \rho_{\text{data}}.
\end{align}
Here, we again use the mean-field assumption, this time for irrelevant variables:
each mask value $\langle m_i \rangle$ is taken to be the ratio between the total number of selected irrelevant variables, $\sum_i m_i - \sum_i s_i$, and the total number of true irrelevant variables, $N - \sum_i s_i$. 

Next, we also derive the selection uncertainty.
In the under-selection case with no false positives, the uncertainty is given by
\begin{align}
    \mathcal{\sigma}_{\text{sel}} &= \frac{1}{N} \sum_{i=1}^N \big \langle  m_i \big \rangle \big \langle 1 - m_i \big \rangle, \nonumber \\
    &= \frac{\sum_i s_i}{N} \frac{\sum_i m_i}{\sum_i s_i} \Big( 1 - \frac{\sum_i m_i}{\sum_i s_i} \Big) \nonumber \\
    &= \frac{\sum_i m_i/N}{\sum_i s_i /N} \Big(\frac{\sum_i s_i}{N} - \frac{\sum_i m_i}{N} \Big) \nonumber \\
    &= \dfrac{\rho_{\text{model}}}{\rho_{\text{data}}} \big(\rho_{\text{data}} - \rho_{\text{model}} \big).
\end{align}
Here, we again apply the mean-field assumption for a relevant variable's mask value, $\langle m_i \rangle = \sum_i m_i / \sum_i s_i$, reflecting that the model selects fewer relevant variables than the true total.
Similarly, in the over-selection case with no false negatives, the uncertainty is
\begin{align}
    \mathcal{\sigma}_{\text{sel}} &= \frac{1}{N} \sum_{i=1}^N \big \langle  m_i \big \rangle \big \langle 1 - m_i \big \rangle, \nonumber \\
    &= \frac{N - \sum_i s_i}{N} \frac{\sum_i m_i - \sum_i s_i}{N-\sum_i s_i} \Big( 1 - \frac{\sum_i m_i - \sum_i s_i}{N -\sum_i s_i} \Big) \nonumber \\
    &= \Big( \frac{\sum_i m_i}{N} - \frac{\sum_i s_i}{N} \Big) \Big(\frac{N - \sum_i m_i}{N - \sum_i s_i} \Big) \nonumber \\
    &= \big(\rho_{\text{model}} - \rho_{\text{data}} \big) \dfrac{1-\rho_{\text{model}}}{1-\rho_{\text{data}}}.
\end{align}
In this over-selection case, we use the mean-field assumption that the mask value for an irrelevant variable is $\langle m_i \rangle = (\sum_i m_i - \sum_i s_i)/(N - \sum_i s_i)$, reflecting that the model selects more relevant variables than the true total.

\section{Inference of data sparsity}
\label{sec:appendix2}
Using the theoretical curve, $\sigma_{\rm sel}(\rho_{\rm model}) = f(\rho_{\rm model}; \rho_{\rm data})$, in Eq.~(\ref{eq:kernel}) for the single-datum density, we infer the true data sparsity $\rho_{\rm data}$ via a relative-fitness mixture fit. 
Specifically, we approximate the observed uncertainty of the selection mask values as a nonnegative linear combination of template curves $f(\rho_{\rm model}; \rho_{\rm data}^{(k)})$ evaluated over a grid of candidate values $\{\rho_{\rm data}^{(k)}\}$:
\begin{equation}
{\sigma}_{\rm sel}(\rho_{\rm model})
= \sum_k p_k\, f(\rho_{\rm model}; \rho_{\rm data}^{(k)}), 
\qquad p_k \ge 0 .
\end{equation}
The normalized coefficients are then interpreted as a discrete posterior distribution:
\begin{equation}
P(\rho_{\rm data}^{(k)}) = \frac{p_k}{\sum_l p_l},
\end{equation}
and the maximizer of this distribution is taken as the point estimate of $\rho_{\rm data}$.

\bibliographystyle{apsrev4-1}
\bibliography{reference}

\begin{thebibliography}{51}%
\makeatletter
\providecommand \@ifxundefined [1]{%
 \@ifx{#1\undefined}
}%
\providecommand \@ifnum [1]{%
 \ifnum #1\expandafter \@firstoftwo
 \else \expandafter \@secondoftwo
 \fi
}%
\providecommand \@ifx [1]{%
 \ifx #1\expandafter \@firstoftwo
 \else \expandafter \@secondoftwo
 \fi
}%
\providecommand \natexlab [1]{#1}%
\providecommand \enquote  [1]{``#1''}%
\providecommand \bibnamefont  [1]{#1}%
\providecommand \bibfnamefont [1]{#1}%
\providecommand \citenamefont [1]{#1}%
\providecommand \href@noop [0]{\@secondoftwo}%
\providecommand \href [0]{\begingroup \@sanitize@url \@href}%
\providecommand \@href[1]{\@@startlink{#1}\@@href}%
\providecommand \@@href[1]{\endgroup#1\@@endlink}%
\providecommand \@sanitize@url [0]{\catcode `\\12\catcode `\$12\catcode `\&12\catcode `\#12\catcode `\^12\catcode `\_12\catcode `\%12\relax}%
\providecommand \@@startlink[1]{}%
\providecommand \@@endlink[0]{}%
\providecommand \url  [0]{\begingroup\@sanitize@url \@url }%
\providecommand \@url [1]{\endgroup\@href {#1}{\urlprefix }}%
\providecommand \urlprefix  [0]{URL }%
\providecommand \Eprint [0]{\href }%
\providecommand \doibase [0]{http://dx.doi.org/}%
\providecommand \selectlanguage [0]{\@gobble}%
\providecommand \bibinfo  [0]{\@secondoftwo}%
\providecommand \bibfield  [0]{\@secondoftwo}%
\providecommand \translation [1]{[#1]}%
\providecommand \BibitemOpen [0]{}%
\providecommand \bibitemStop [0]{}%
\providecommand \bibitemNoStop [0]{.\EOS\space}%
\providecommand \EOS [0]{\spacefactor3000\relax}%
\providecommand \BibitemShut  [1]{\csname bibitem#1\endcsname}%
\let\auto@bib@innerbib\@empty
%</preamble>
\bibitem [{\citenamefont {Chatterjee}\ and\ \citenamefont {Hadi}(2015)}]{chatterjee2015regression}%
  \BibitemOpen
  \bibfield  {author} {\bibinfo {author} {\bibfnamefont {S.}~\bibnamefont {Chatterjee}}\ and\ \bibinfo {author} {\bibfnamefont {A.~S.}\ \bibnamefont {Hadi}},\ }\href@noop {} {\emph {\bibinfo {title} {Regression analysis by example}}}\ (\bibinfo  {publisher} {John Wiley \& Sons},\ \bibinfo {year} {2015})\BibitemShut {NoStop}%
\bibitem [{\citenamefont {James}\ \emph {et~al.}(2023)\citenamefont {James}, \citenamefont {Witten}, \citenamefont {Hastie}, \citenamefont {Tibshirani},\ and\ \citenamefont {Taylor}}]{james2023linear}%
  \BibitemOpen
  \bibfield  {author} {\bibinfo {author} {\bibfnamefont {G.}~\bibnamefont {James}}, \bibinfo {author} {\bibfnamefont {D.}~\bibnamefont {Witten}}, \bibinfo {author} {\bibfnamefont {T.}~\bibnamefont {Hastie}}, \bibinfo {author} {\bibfnamefont {R.}~\bibnamefont {Tibshirani}}, \ and\ \bibinfo {author} {\bibfnamefont {J.}~\bibnamefont {Taylor}},\ }in\ \href@noop {} {\emph {\bibinfo {booktitle} {An introduction to statistical learning: With applications in}}}\ (\bibinfo  {publisher} {Springer},\ \bibinfo {year} {2023})\ pp.\ \bibinfo {pages} {69--134}\BibitemShut {NoStop}%
\bibitem [{\citenamefont {Majda}\ and\ \citenamefont {Harlim}(2012)}]{majda2012physics}%
  \BibitemOpen
  \bibfield  {author} {\bibinfo {author} {\bibfnamefont {A.~J.}\ \bibnamefont {Majda}}\ and\ \bibinfo {author} {\bibfnamefont {J.}~\bibnamefont {Harlim}},\ }\href@noop {} {\bibfield  {journal} {\bibinfo  {journal} {Nonlinearity}\ }\textbf {\bibinfo {volume} {26}},\ \bibinfo {pages} {201} (\bibinfo {year} {2012})}\BibitemShut {NoStop}%
\bibitem [{\citenamefont {Kaptanoglu}\ \emph {et~al.}(2023)\citenamefont {Kaptanoglu}, \citenamefont {Hansen}, \citenamefont {Lore}, \citenamefont {Landreman},\ and\ \citenamefont {Brunton}}]{kaptanoglu2023sparse}%
  \BibitemOpen
  \bibfield  {author} {\bibinfo {author} {\bibfnamefont {A.~A.}\ \bibnamefont {Kaptanoglu}}, \bibinfo {author} {\bibfnamefont {C.}~\bibnamefont {Hansen}}, \bibinfo {author} {\bibfnamefont {J.~D.}\ \bibnamefont {Lore}}, \bibinfo {author} {\bibfnamefont {M.}~\bibnamefont {Landreman}}, \ and\ \bibinfo {author} {\bibfnamefont {S.~L.}\ \bibnamefont {Brunton}},\ }\href@noop {} {\bibfield  {journal} {\bibinfo  {journal} {Phys. Plasmas}\ }\textbf {\bibinfo {volume} {30}} (\bibinfo {year} {2023})}\BibitemShut {NoStop}%
\bibitem [{\citenamefont {Brunton}\ \emph {et~al.}(2016)\citenamefont {Brunton}, \citenamefont {Proctor},\ and\ \citenamefont {Kutz}}]{Brunton2016}%
  \BibitemOpen
  \bibfield  {author} {\bibinfo {author} {\bibfnamefont {S.~L.}\ \bibnamefont {Brunton}}, \bibinfo {author} {\bibfnamefont {J.~L.}\ \bibnamefont {Proctor}}, \ and\ \bibinfo {author} {\bibfnamefont {J.~N.}\ \bibnamefont {Kutz}},\ }\href {\doibase 10.1073/pnas.1517384113} {\bibfield  {journal} {\bibinfo  {journal} {Proc. Natl. Acad. Sci. U.S.A.}\ }\textbf {\bibinfo {volume} {113}},\ \bibinfo {pages} {3932} (\bibinfo {year} {2016})}\BibitemShut {NoStop}%
\bibitem [{\citenamefont {Quade}\ \emph {et~al.}(2016)\citenamefont {Quade}, \citenamefont {Abel}, \citenamefont {Shafi}, \citenamefont {Niven},\ and\ \citenamefont {Noack}}]{Quade2016}%
  \BibitemOpen
  \bibfield  {author} {\bibinfo {author} {\bibfnamefont {M.}~\bibnamefont {Quade}}, \bibinfo {author} {\bibfnamefont {M.}~\bibnamefont {Abel}}, \bibinfo {author} {\bibfnamefont {K.}~\bibnamefont {Shafi}}, \bibinfo {author} {\bibfnamefont {R.~K.}\ \bibnamefont {Niven}}, \ and\ \bibinfo {author} {\bibfnamefont {B.~R.}\ \bibnamefont {Noack}},\ }\href {\doibase 10.1103/PhysRevE.94.012214} {\bibfield  {journal} {\bibinfo  {journal} {Phys. Rev. E}\ }\textbf {\bibinfo {volume} {94}},\ \bibinfo {pages} {012214} (\bibinfo {year} {2016})}\BibitemShut {NoStop}%
\bibitem [{\citenamefont {Maalouf}(2011)}]{maalouf2011logistic}%
  \BibitemOpen
  \bibfield  {author} {\bibinfo {author} {\bibfnamefont {M.}~\bibnamefont {Maalouf}},\ }\href@noop {} {\bibfield  {journal} {\bibinfo  {journal} {International Journal of Data Analysis Techniques and Strategies}\ }\textbf {\bibinfo {volume} {3}},\ \bibinfo {pages} {281} (\bibinfo {year} {2011})}\BibitemShut {NoStop}%
\bibitem [{\citenamefont {Zhang}(2004)}]{zhang2004solving}%
  \BibitemOpen
  \bibfield  {author} {\bibinfo {author} {\bibfnamefont {T.}~\bibnamefont {Zhang}},\ }in\ \href@noop {} {\emph {\bibinfo {booktitle} {Proceedings of the twenty-first international conference on Machine learning}}}\ (\bibinfo {year} {2004})\ p.\ \bibinfo {pages} {116}\BibitemShut {NoStop}%
\bibitem [{\citenamefont {Dawes}\ and\ \citenamefont {Corrigan}(1974)}]{dawes1974linear}%
  \BibitemOpen
  \bibfield  {author} {\bibinfo {author} {\bibfnamefont {R.~M.}\ \bibnamefont {Dawes}}\ and\ \bibinfo {author} {\bibfnamefont {B.}~\bibnamefont {Corrigan}},\ }\href@noop {} {\bibfield  {journal} {\bibinfo  {journal} {Psychol. Bull.}\ }\textbf {\bibinfo {volume} {81}},\ \bibinfo {pages} {95} (\bibinfo {year} {1974})}\BibitemShut {NoStop}%
\bibitem [{\citenamefont {Montgomery}\ \emph {et~al.}(2021)\citenamefont {Montgomery}, \citenamefont {Peck},\ and\ \citenamefont {Vining}}]{montgomery2021introduction}%
  \BibitemOpen
  \bibfield  {author} {\bibinfo {author} {\bibfnamefont {D.~C.}\ \bibnamefont {Montgomery}}, \bibinfo {author} {\bibfnamefont {E.~A.}\ \bibnamefont {Peck}}, \ and\ \bibinfo {author} {\bibfnamefont {G.~G.}\ \bibnamefont {Vining}},\ }\href@noop {} {\emph {\bibinfo {title} {Introduction to linear regression analysis}}}\ (\bibinfo  {publisher} {John Wiley \& Sons},\ \bibinfo {year} {2021})\BibitemShut {NoStop}%
\bibitem [{\citenamefont {Cramer}(2002)}]{Cramer2002}%
  \BibitemOpen
  \bibfield  {author} {\bibinfo {author} {\bibfnamefont {J.}~\bibnamefont {Cramer}},\ }\href {\doibase 10.2139/ssrn.360300} {\bibfield  {journal} {\bibinfo  {journal} {Tinbergen Institute, Tinbergen Institute Discussion Papers}\ } (\bibinfo {year} {2002}),\ 10.2139/ssrn.360300}\BibitemShut {NoStop}%
\bibitem [{\citenamefont {Cortes}\ and\ \citenamefont {Vapnik}(1995)}]{Cortes1995}%
  \BibitemOpen
  \bibfield  {author} {\bibinfo {author} {\bibfnamefont {C.}~\bibnamefont {Cortes}}\ and\ \bibinfo {author} {\bibfnamefont {V.}~\bibnamefont {Vapnik}},\ }\href {\doibase 10.1007/BF00994018} {\bibfield  {journal} {\bibinfo  {journal} {Mach. Learn.}\ }\textbf {\bibinfo {volume} {20}},\ \bibinfo {pages} {273} (\bibinfo {year} {1995})}\BibitemShut {NoStop}%
\bibitem [{\citenamefont {Nelder}\ and\ \citenamefont {Wedderburn}(1972)}]{Nelder1972}%
  \BibitemOpen
  \bibfield  {author} {\bibinfo {author} {\bibfnamefont {J.~A.}\ \bibnamefont {Nelder}}\ and\ \bibinfo {author} {\bibfnamefont {R.~W.~M.}\ \bibnamefont {Wedderburn}},\ }\href {http://www.jstor.org/stable/2344614} {\bibfield  {journal} {\bibinfo  {journal} {J. R. Stat. Soc. A}\ }\textbf {\bibinfo {volume} {135}},\ \bibinfo {pages} {370} (\bibinfo {year} {1972})}\BibitemShut {NoStop}%
\bibitem [{\citenamefont {Zhou}\ \emph {et~al.}(2024)\citenamefont {Zhou}, \citenamefont {Yan},\ and\ \citenamefont {Zhang}}]{Zhou2024}%
  \BibitemOpen
  \bibfield  {author} {\bibinfo {author} {\bibfnamefont {W.}~\bibnamefont {Zhou}}, \bibinfo {author} {\bibfnamefont {Z.}~\bibnamefont {Yan}}, \ and\ \bibinfo {author} {\bibfnamefont {L.}~\bibnamefont {Zhang}},\ }\href {\doibase 10.1038/s41598-024-55243-x} {\bibfield  {journal} {\bibinfo  {journal} {Sci. Rep.}\ }\textbf {\bibinfo {volume} {14}},\ \bibinfo {pages} {5905} (\bibinfo {year} {2024})}\BibitemShut {NoStop}%
\bibitem [{\citenamefont {Shastry}(2007)}]{Shastry2007}%
  \BibitemOpen
  \bibfield  {author} {\bibinfo {author} {\bibfnamefont {B.~S.}\ \bibnamefont {Shastry}},\ }\href {\doibase 10.1007/s10038-007-0200-z} {\bibfield  {journal} {\bibinfo  {journal} {J. Hum. Genet.}\ }\textbf {\bibinfo {volume} {52}},\ \bibinfo {pages} {871} (\bibinfo {year} {2007})}\BibitemShut {NoStop}%
\bibitem [{\citenamefont {Olsson}\ \emph {et~al.}(2016)\citenamefont {Olsson}, \citenamefont {Lautner}, \citenamefont {Andreasson}, \citenamefont {{\"O}hrfelt}, \citenamefont {Portelius}, \citenamefont {Bjerke}, \citenamefont {H{\"o}ltt{\"a}}, \citenamefont {Ros{\'e}n}, \citenamefont {Olsson}, \citenamefont {Strobel}, \citenamefont {Wu}, \citenamefont {Dakin}, \citenamefont {Petzold}, \citenamefont {Blennow},\ and\ \citenamefont {Zetterberg}}]{Olsson2016}%
  \BibitemOpen
  \bibfield  {author} {\bibinfo {author} {\bibfnamefont {B.}~\bibnamefont {Olsson}}, \bibinfo {author} {\bibfnamefont {R.}~\bibnamefont {Lautner}}, \bibinfo {author} {\bibfnamefont {U.}~\bibnamefont {Andreasson}}, \bibinfo {author} {\bibfnamefont {A.}~\bibnamefont {{\"O}hrfelt}}, \bibinfo {author} {\bibfnamefont {E.}~\bibnamefont {Portelius}}, \bibinfo {author} {\bibfnamefont {M.}~\bibnamefont {Bjerke}}, \bibinfo {author} {\bibfnamefont {M.}~\bibnamefont {H{\"o}ltt{\"a}}}, \bibinfo {author} {\bibfnamefont {C.}~\bibnamefont {Ros{\'e}n}}, \bibinfo {author} {\bibfnamefont {C.}~\bibnamefont {Olsson}}, \bibinfo {author} {\bibfnamefont {G.}~\bibnamefont {Strobel}}, \bibinfo {author} {\bibfnamefont {E.}~\bibnamefont {Wu}}, \bibinfo {author} {\bibfnamefont {K.}~\bibnamefont {Dakin}}, \bibinfo {author} {\bibfnamefont {M.}~\bibnamefont {Petzold}}, \bibinfo {author} {\bibfnamefont {K.}~\bibnamefont {Blennow}}, \ and\ \bibinfo {author} {\bibfnamefont {H.}~\bibnamefont {Zetterberg}},\ }\href@noop {} {\bibfield  {journal}
  {\bibinfo  {journal} {Lancet Neurol.}\ }\textbf {\bibinfo {volume} {15}},\ \bibinfo {pages} {673} (\bibinfo {year} {2016})}\BibitemShut {NoStop}%
\bibitem [{\citenamefont {Einav}\ and\ \citenamefont {Levin}(2014)}]{Einav2014}%
  \BibitemOpen
  \bibfield  {author} {\bibinfo {author} {\bibfnamefont {L.}~\bibnamefont {Einav}}\ and\ \bibinfo {author} {\bibfnamefont {J.}~\bibnamefont {Levin}},\ }\href {\doibase 10.1126/science.1243089} {\bibfield  {journal} {\bibinfo  {journal} {Science}\ }\textbf {\bibinfo {volume} {346}},\ \bibinfo {pages} {1243089} (\bibinfo {year} {2014})}\BibitemShut {NoStop}%
\bibitem [{\citenamefont {Han}\ \emph {et~al.}(2015)\citenamefont {Han}, \citenamefont {Pool}, \citenamefont {Tran},\ and\ \citenamefont {Dally}}]{Han2015}%
  \BibitemOpen
  \bibfield  {author} {\bibinfo {author} {\bibfnamefont {S.}~\bibnamefont {Han}}, \bibinfo {author} {\bibfnamefont {J.}~\bibnamefont {Pool}}, \bibinfo {author} {\bibfnamefont {J.}~\bibnamefont {Tran}}, \ and\ \bibinfo {author} {\bibfnamefont {W.~J.}\ \bibnamefont {Dally}},\ }in\ \href@noop {} {\emph {\bibinfo {booktitle} {Proceedings of the 29th International Conference on Neural Information Processing Systems - Volume 1}}},\ \bibinfo {series and number} {NIPS'15}\ (\bibinfo  {publisher} {MIT Press},\ \bibinfo {address} {Cambridge, MA, USA},\ \bibinfo {year} {2015})\ p.\ \bibinfo {pages} {1135–1143}\BibitemShut {NoStop}%
\bibitem [{\citenamefont {Frankle}\ and\ \citenamefont {Carbin}(2019)}]{Frankle2018}%
  \BibitemOpen
  \bibfield  {author} {\bibinfo {author} {\bibfnamefont {J.}~\bibnamefont {Frankle}}\ and\ \bibinfo {author} {\bibfnamefont {M.}~\bibnamefont {Carbin}},\ }in\ \href {https://openreview.net/forum?id=rJl-b3RcF7} {\emph {\bibinfo {booktitle} {International Conference on Learning Representations}}}\ (\bibinfo {year} {2019})\BibitemShut {NoStop}%
\bibitem [{\citenamefont {Clark}\ \emph {et~al.}(2019)\citenamefont {Clark}, \citenamefont {Khandelwal}, \citenamefont {Levy},\ and\ \citenamefont {Manning}}]{Clark2019}%
  \BibitemOpen
  \bibfield  {author} {\bibinfo {author} {\bibfnamefont {K.}~\bibnamefont {Clark}}, \bibinfo {author} {\bibfnamefont {U.}~\bibnamefont {Khandelwal}}, \bibinfo {author} {\bibfnamefont {O.}~\bibnamefont {Levy}}, \ and\ \bibinfo {author} {\bibfnamefont {C.~D.}\ \bibnamefont {Manning}},\ }in\ \href {\doibase 10.18653/v1/W19-4828} {\emph {\bibinfo {booktitle} {Proceedings of the 2019 ACL Workshop BlackboxNLP: Analyzing and Interpreting Neural Networks for NLP}}},\ \bibinfo {editor} {edited by\ \bibinfo {editor} {\bibfnamefont {T.}~\bibnamefont {Linzen}}, \bibinfo {editor} {\bibfnamefont {G.}~\bibnamefont {Chrupa{\l}a}}, \bibinfo {editor} {\bibfnamefont {Y.}~\bibnamefont {Belinkov}}, \ and\ \bibinfo {editor} {\bibfnamefont {D.}~\bibnamefont {Hupkes}}}\ (\bibinfo  {publisher} {Association for Computational Linguistics},\ \bibinfo {address} {Florence, Italy},\ \bibinfo {year} {2019})\ pp.\ \bibinfo {pages} {276--286}\BibitemShut {NoStop}%
\bibitem [{\citenamefont {Michel}\ \emph {et~al.}(2019)\citenamefont {Michel}, \citenamefont {Levy},\ and\ \citenamefont {Neubig}}]{Michel2019}%
  \BibitemOpen
  \bibfield  {author} {\bibinfo {author} {\bibfnamefont {P.}~\bibnamefont {Michel}}, \bibinfo {author} {\bibfnamefont {O.}~\bibnamefont {Levy}}, \ and\ \bibinfo {author} {\bibfnamefont {G.}~\bibnamefont {Neubig}},\ }in\ \href {https://proceedings.neurips.cc/paper_files/paper/2019/file/2c601ad9d2ff9bc8b282670cdd54f69f-Paper.pdf} {\emph {\bibinfo {booktitle} {Advances in Neural Information Processing Systems}}},\ Vol.~\bibinfo {volume} {32},\ \bibinfo {editor} {edited by\ \bibinfo {editor} {\bibfnamefont {H.}~\bibnamefont {Wallach}}, \bibinfo {editor} {\bibfnamefont {H.}~\bibnamefont {Larochelle}}, \bibinfo {editor} {\bibfnamefont {A.}~\bibnamefont {Beygelzimer}}, \bibinfo {editor} {\bibfnamefont {F.}~\bibnamefont {d\textquotesingle Alch\'{e}-Buc}}, \bibinfo {editor} {\bibfnamefont {E.}~\bibnamefont {Fox}}, \ and\ \bibinfo {editor} {\bibfnamefont {R.}~\bibnamefont {Garnett}}}\ (\bibinfo  {publisher} {Curran Associates, Inc.},\ \bibinfo {year} {2019})\BibitemShut {NoStop}%
\bibitem [{\citenamefont {Babyak}(2004)}]{babyak2004you}%
  \BibitemOpen
  \bibfield  {author} {\bibinfo {author} {\bibfnamefont {M.~A.}\ \bibnamefont {Babyak}},\ }\href@noop {} {\bibfield  {journal} {\bibinfo  {journal} {Biopsychosocial Science and Medicine}\ }\textbf {\bibinfo {volume} {66}},\ \bibinfo {pages} {411} (\bibinfo {year} {2004})}\BibitemShut {NoStop}%
\bibitem [{\citenamefont {Hawkins}(2004)}]{hawkins2004problem}%
  \BibitemOpen
  \bibfield  {author} {\bibinfo {author} {\bibfnamefont {D.~M.}\ \bibnamefont {Hawkins}},\ }\href@noop {} {\bibfield  {journal} {\bibinfo  {journal} {J. Chem. Inf. Comput. Sci.}\ }\textbf {\bibinfo {volume} {44}},\ \bibinfo {pages} {1} (\bibinfo {year} {2004})}\BibitemShut {NoStop}%
\bibitem [{\citenamefont {Gruber}(1998)}]{ref:BiasVariance}%
  \BibitemOpen
  \bibfield  {author} {\bibinfo {author} {\bibfnamefont {M.}~\bibnamefont {Gruber}},\ }\href {\doibase 10.1201/9780203751220} {\emph {\bibinfo {title} {Improving Efficiency by Shrinkage: The James--Stein and Ridge Regression Estimators}}},\ \bibinfo {edition} {1st}\ ed.\ (\bibinfo  {publisher} {Routledge},\ \bibinfo {year} {1998})\BibitemShut {NoStop}%
\bibitem [{\citenamefont {Hoerl}\ and\ \citenamefont {Kennard}(1970)}]{ref:Ridge}%
  \BibitemOpen
  \bibfield  {author} {\bibinfo {author} {\bibfnamefont {A.~E.}\ \bibnamefont {Hoerl}}\ and\ \bibinfo {author} {\bibfnamefont {R.~W.}\ \bibnamefont {Kennard}},\ }\href {\doibase 10.1080/00401706.1970.10488634} {\bibfield  {journal} {\bibinfo  {journal} {Technometrics}\ }\textbf {\bibinfo {volume} {12}},\ \bibinfo {pages} {55} (\bibinfo {year} {1970})}\BibitemShut {NoStop}%
\bibitem [{\citenamefont {Tibshirani}(1996)}]{ref:LASSO}%
  \BibitemOpen
  \bibfield  {author} {\bibinfo {author} {\bibfnamefont {R.}~\bibnamefont {Tibshirani}},\ }\href {http://www.jstor.org/stable/2346178} {\bibfield  {journal} {\bibinfo  {journal} {J. R. Stat. Soc. B (Methodol.)}\ }\textbf {\bibinfo {volume} {58}},\ \bibinfo {pages} {267} (\bibinfo {year} {1996})}\BibitemShut {NoStop}%
\bibitem [{\citenamefont {Ng}(2011)}]{ng2011sparse}%
  \BibitemOpen
  \bibfield  {author} {\bibinfo {author} {\bibfnamefont {A.}~\bibnamefont {Ng}},\ }\href@noop {} {\enquote {\bibinfo {title} {Sparse autoencoder},}\ } (\bibinfo {year} {2011}),\ \bibinfo {note} {unpublished lecture notes.}\BibitemShut {Stop}%
\bibitem [{\citenamefont {Huben}\ \emph {et~al.}(2024)\citenamefont {Huben}, \citenamefont {Cunningham}, \citenamefont {Smith}, \citenamefont {Ewart},\ and\ \citenamefont {Sharkey}}]{huben2024sparse}%
  \BibitemOpen
  \bibfield  {author} {\bibinfo {author} {\bibfnamefont {R.}~\bibnamefont {Huben}}, \bibinfo {author} {\bibfnamefont {H.}~\bibnamefont {Cunningham}}, \bibinfo {author} {\bibfnamefont {L.~R.}\ \bibnamefont {Smith}}, \bibinfo {author} {\bibfnamefont {A.}~\bibnamefont {Ewart}}, \ and\ \bibinfo {author} {\bibfnamefont {L.}~\bibnamefont {Sharkey}},\ }in\ \href {https://openreview.net/forum?id=F76bwRSLeK} {\emph {\bibinfo {booktitle} {The Twelfth International Conference on Learning Representations}}}\ (\bibinfo {year} {2024})\BibitemShut {NoStop}%
\bibitem [{\citenamefont {Breiman}(1995)}]{ref:NNGarrote}%
  \BibitemOpen
  \bibfield  {author} {\bibinfo {author} {\bibfnamefont {L.}~\bibnamefont {Breiman}},\ }\href {http://www.jstor.org/stable/1269730} {\bibfield  {journal} {\bibinfo  {journal} {Technometrics}\ }\textbf {\bibinfo {volume} {37}},\ \bibinfo {pages} {373} (\bibinfo {year} {1995})}\BibitemShut {NoStop}%
\bibitem [{\citenamefont {Kappen}\ and\ \citenamefont {G{\'o}mez}(2014)}]{ref:VG}%
  \BibitemOpen
  \bibfield  {author} {\bibinfo {author} {\bibfnamefont {H.~J.}\ \bibnamefont {Kappen}}\ and\ \bibinfo {author} {\bibfnamefont {V.}~\bibnamefont {G{\'o}mez}},\ }\href {\doibase 10.1007/s10994-013-5427-7} {\bibfield  {journal} {\bibinfo  {journal} {Mach. Learn.}\ }\textbf {\bibinfo {volume} {96}},\ \bibinfo {pages} {269} (\bibinfo {year} {2014})}\BibitemShut {NoStop}%
\bibitem [{\citenamefont {Penrose}(1955)}]{Penrose1955}%
  \BibitemOpen
  \bibfield  {author} {\bibinfo {author} {\bibfnamefont {R.}~\bibnamefont {Penrose}},\ }\href {\doibase 10.1017/S0305004100030401} {\bibfield  {journal} {\bibinfo  {journal} {Math. Proc. Cambridge Philos. Soc.}\ }\textbf {\bibinfo {volume} {51}},\ \bibinfo {pages} {406–413} (\bibinfo {year} {1955})}\BibitemShut {NoStop}%
\bibitem [{\citenamefont {Beck}\ and\ \citenamefont {Teboulle}(2009)}]{BeckTeboulle2009}%
  \BibitemOpen
  \bibfield  {author} {\bibinfo {author} {\bibfnamefont {A.}~\bibnamefont {Beck}}\ and\ \bibinfo {author} {\bibfnamefont {M.}~\bibnamefont {Teboulle}},\ }\href@noop {} {\bibfield  {journal} {\bibinfo  {journal} {SIAM J. Imaging Sci.}\ }\textbf {\bibinfo {volume} {2}},\ \bibinfo {pages} {183} (\bibinfo {year} {2009})}\BibitemShut {NoStop}%
\bibitem [{\citenamefont {Friedman}\ \emph {et~al.}(2010)\citenamefont {Friedman}, \citenamefont {Hastie},\ and\ \citenamefont {Tibshirani}}]{Friedman2010}%
  \BibitemOpen
  \bibfield  {author} {\bibinfo {author} {\bibfnamefont {J.}~\bibnamefont {Friedman}}, \bibinfo {author} {\bibfnamefont {T.}~\bibnamefont {Hastie}}, \ and\ \bibinfo {author} {\bibfnamefont {R.}~\bibnamefont {Tibshirani}},\ }\href@noop {} {\bibfield  {journal} {\bibinfo  {journal} {J. Stat. Softw.}\ }\textbf {\bibinfo {volume} {33}},\ \bibinfo {pages} {1} (\bibinfo {year} {2010})}\BibitemShut {NoStop}%
\bibitem [{\citenamefont {Boyd}\ \emph {et~al.}(2011)\citenamefont {Boyd}, \citenamefont {Parikh}, \citenamefont {Chu}, \citenamefont {Peleato},\ and\ \citenamefont {Eckstein}}]{Boyd2011}%
  \BibitemOpen
  \bibfield  {author} {\bibinfo {author} {\bibfnamefont {S.}~\bibnamefont {Boyd}}, \bibinfo {author} {\bibfnamefont {N.}~\bibnamefont {Parikh}}, \bibinfo {author} {\bibfnamefont {E.}~\bibnamefont {Chu}}, \bibinfo {author} {\bibfnamefont {B.}~\bibnamefont {Peleato}}, \ and\ \bibinfo {author} {\bibfnamefont {J.}~\bibnamefont {Eckstein}},\ }\href@noop {} {\bibfield  {journal} {\bibinfo  {journal} {Found. Trends Mach. Learn.}\ }\textbf {\bibinfo {volume} {3}},\ \bibinfo {pages} {1} (\bibinfo {year} {2011})}\BibitemShut {NoStop}%
\bibitem [{\citenamefont {Zou}\ and\ \citenamefont {Hastie}(2005)}]{ref:ElasticNet}%
  \BibitemOpen
  \bibfield  {author} {\bibinfo {author} {\bibfnamefont {H.}~\bibnamefont {Zou}}\ and\ \bibinfo {author} {\bibfnamefont {T.}~\bibnamefont {Hastie}},\ }\href {\doibase 10.1111/j.1467-9868.2005.00503.x} {\bibfield  {journal} {\bibinfo  {journal} {J. R. Stat. Soc. B}\ }\textbf {\bibinfo {volume} {67}},\ \bibinfo {pages} {301} (\bibinfo {year} {2005})}\BibitemShut {NoStop}%
\bibitem [{\citenamefont {Mitchell}\ and\ \citenamefont {Beauchamp}(1988)}]{ref:SpikeAndSlab}%
  \BibitemOpen
  \bibfield  {author} {\bibinfo {author} {\bibfnamefont {T.~J.}\ \bibnamefont {Mitchell}}\ and\ \bibinfo {author} {\bibfnamefont {J.~J.}\ \bibnamefont {Beauchamp}},\ }\href {\doibase 10.1080/01621459.1988.10478694} {\bibfield  {journal} {\bibinfo  {journal} {J. Am. Stat. Assoc.}\ }\textbf {\bibinfo {volume} {83}},\ \bibinfo {pages} {1023} (\bibinfo {year} {1988})}\BibitemShut {NoStop}%
\bibitem [{\citenamefont {George}\ and\ \citenamefont {McCulloch}(1997)}]{ref:BayesSpSl}%
  \BibitemOpen
  \bibfield  {author} {\bibinfo {author} {\bibfnamefont {E.~I.}\ \bibnamefont {George}}\ and\ \bibinfo {author} {\bibfnamefont {R.~E.}\ \bibnamefont {McCulloch}},\ }\href {http://www.jstor.org/stable/24306083} {\bibfield  {journal} {\bibinfo  {journal} {Stat. Sin.}\ }\textbf {\bibinfo {volume} {7}},\ \bibinfo {pages} {339} (\bibinfo {year} {1997})}\BibitemShut {NoStop}%
\bibitem [{\citenamefont {M{\'e}zard}\ \emph {et~al.}(1987)\citenamefont {M{\'e}zard}, \citenamefont {Parisi},\ and\ \citenamefont {Virasoro}}]{mezard1987spin}%
  \BibitemOpen
  \bibfield  {author} {\bibinfo {author} {\bibfnamefont {M.}~\bibnamefont {M{\'e}zard}}, \bibinfo {author} {\bibfnamefont {G.}~\bibnamefont {Parisi}}, \ and\ \bibinfo {author} {\bibfnamefont {M.~A.}\ \bibnamefont {Virasoro}},\ }\href@noop {} {\emph {\bibinfo {title} {Spin Glass Theory and Beyond}}}\ (\bibinfo  {publisher} {World Scientific},\ \bibinfo {year} {1987})\BibitemShut {NoStop}%
\bibitem [{\citenamefont {Zhang}\ and\ \citenamefont {Huang}(2008)}]{ee685348-c1b0-32a8-849d-4be4de07f8ad}%
  \BibitemOpen
  \bibfield  {author} {\bibinfo {author} {\bibfnamefont {C.-H.}\ \bibnamefont {Zhang}}\ and\ \bibinfo {author} {\bibfnamefont {J.}~\bibnamefont {Huang}},\ }\href {http://www.jstor.org/stable/25464684} {\bibfield  {journal} {\bibinfo  {journal} {The Annals of Statistics}\ }\textbf {\bibinfo {volume} {36}},\ \bibinfo {pages} {1567} (\bibinfo {year} {2008})}\BibitemShut {NoStop}%
\bibitem [{\citenamefont {Zhou}(2010)}]{Zhou2010ThresholdedLF}%
  \BibitemOpen
  \bibfield  {author} {\bibinfo {author} {\bibfnamefont {S.}~\bibnamefont {Zhou}},\ }\href {https://api.semanticscholar.org/CorpusID:5341602} {\bibfield  {journal} {\bibinfo  {journal} {arXiv: Statistics Theory}\ } (\bibinfo {year} {2010})}\BibitemShut {NoStop}%
\bibitem [{\citenamefont {Redmond}(2002)}]{ref:CCDataset}%
  \BibitemOpen
  \bibfield  {author} {\bibinfo {author} {\bibfnamefont {M.}~\bibnamefont {Redmond}},\ }\href@noop {} {\enquote {\bibinfo {title} {{Communities and Crime}},}\ }\bibinfo {howpublished} {UCI Machine Learning Repository} (\bibinfo {year} {2002}),\ \bibinfo {note} {{DOI}: https://doi.org/10.24432/C53W3X}\BibitemShut {NoStop}%
\bibitem [{\citenamefont {Buza}(2014)}]{ref:BlogFeedback}%
  \BibitemOpen
  \bibfield  {author} {\bibinfo {author} {\bibfnamefont {K.}~\bibnamefont {Buza}},\ }\href@noop {} {\enquote {\bibinfo {title} {{BlogFeedback}},}\ }\bibinfo {howpublished} {UCI Machine Learning Repository} (\bibinfo {year} {2014}),\ \bibinfo {note} {{DOI}: https://doi.org/10.24432/C58S3F}\BibitemShut {NoStop}%
\bibitem [{\citenamefont {Hansen}\ \emph {et~al.}(2013)\citenamefont {Hansen}, \citenamefont {Stahlhut},\ and\ \citenamefont {Hansen}}]{HansenStahlhutHansen2013}%
  \BibitemOpen
  \bibfield  {author} {\bibinfo {author} {\bibfnamefont {S.~T.}\ \bibnamefont {Hansen}}, \bibinfo {author} {\bibfnamefont {C.}~\bibnamefont {Stahlhut}}, \ and\ \bibinfo {author} {\bibfnamefont {L.~K.}\ \bibnamefont {Hansen}},\ }in\ \href {\doibase 10.1109/PRNI.2013.36} {\emph {\bibinfo {booktitle} {Proc.\ 2013 Int.\ Workshop on Pattern Recognition in Neuroimaging (PRNI)}}}\ (\bibinfo  {publisher} {IEEE},\ \bibinfo {year} {2013})\ pp.\ \bibinfo {pages} {106--109}\BibitemShut {NoStop}%
\bibitem [{\citenamefont {Andersen}\ \emph {et~al.}(2013)\citenamefont {Andersen}, \citenamefont {Hansen},\ and\ \citenamefont {Hansen}}]{AndersenHansenHansen2013}%
  \BibitemOpen
  \bibfield  {author} {\bibinfo {author} {\bibfnamefont {M.~R.}\ \bibnamefont {Andersen}}, \bibinfo {author} {\bibfnamefont {S.~T.}\ \bibnamefont {Hansen}}, \ and\ \bibinfo {author} {\bibfnamefont {L.~K.}\ \bibnamefont {Hansen}},\ }in\ \href {\doibase 10.1109/MLSP.2013.6661919} {\emph {\bibinfo {booktitle} {2013 IEEE Int.\ Workshop on Machine Learning for Signal Processing (MLSP)}}}\ (\bibinfo  {publisher} {IEEE},\ \bibinfo {year} {2013})\ pp.\ \bibinfo {pages} {1--6}\BibitemShut {NoStop}%
\bibitem [{\citenamefont {Donoho}(2006)}]{Donoho2006}%
  \BibitemOpen
  \bibfield  {author} {\bibinfo {author} {\bibfnamefont {D.~L.}\ \bibnamefont {Donoho}},\ }\href@noop {} {\bibfield  {journal} {\bibinfo  {journal} {IEEE Trans. Inf. Theory}\ }\textbf {\bibinfo {volume} {52}},\ \bibinfo {pages} {1289} (\bibinfo {year} {2006})}\BibitemShut {NoStop}%
\bibitem [{\citenamefont {Lustig}\ \emph {et~al.}(2007)\citenamefont {Lustig}, \citenamefont {Donoho},\ and\ \citenamefont {Pauly}}]{Lustig2007}%
  \BibitemOpen
  \bibfield  {author} {\bibinfo {author} {\bibfnamefont {M.}~\bibnamefont {Lustig}}, \bibinfo {author} {\bibfnamefont {D.}~\bibnamefont {Donoho}}, \ and\ \bibinfo {author} {\bibfnamefont {J.~M.}\ \bibnamefont {Pauly}},\ }\href@noop {} {\bibfield  {journal} {\bibinfo  {journal} {Magn. Reson. Med.}\ }\textbf {\bibinfo {volume} {58}},\ \bibinfo {pages} {1182} (\bibinfo {year} {2007})}\BibitemShut {NoStop}%
\bibitem [{\citenamefont {Martin}\ and\ \citenamefont {Mahoney}(2021)}]{Martin2021}%
  \BibitemOpen
  \bibfield  {author} {\bibinfo {author} {\bibfnamefont {C.~H.}\ \bibnamefont {Martin}}\ and\ \bibinfo {author} {\bibfnamefont {M.~W.}\ \bibnamefont {Mahoney}},\ }\href {http://jmlr.org/papers/v22/20-410.html} {\bibfield  {journal} {\bibinfo  {journal} {J. Mach. Learn. Res.}\ }\textbf {\bibinfo {volume} {22}},\ \bibinfo {pages} {1} (\bibinfo {year} {2021})}\BibitemShut {NoStop}%
\bibitem [{\citenamefont {Martin}\ and\ \citenamefont {Mahoney}(2020)}]{MartinMahoney2020}%
  \BibitemOpen
  \bibfield  {author} {\bibinfo {author} {\bibfnamefont {C.~H.}\ \bibnamefont {Martin}}\ and\ \bibinfo {author} {\bibfnamefont {M.~W.}\ \bibnamefont {Mahoney}},\ }in\ \href {\doibase 10.1137/1.9781611976236.57} {\emph {\bibinfo {booktitle} {Proceedings of the 2020 SIAM International Conference on Data Mining (SDM)}}}\ (\bibinfo {year} {2020})\ pp.\ \bibinfo {pages} {505--513}\BibitemShut {NoStop}%
\bibitem [{\citenamefont {Meng}\ and\ \citenamefont {Yao}(2023)}]{MengYao2023}%
  \BibitemOpen
  \bibfield  {author} {\bibinfo {author} {\bibfnamefont {X.}~\bibnamefont {Meng}}\ and\ \bibinfo {author} {\bibfnamefont {J.}~\bibnamefont {Yao}},\ }\href@noop {} {\bibfield  {journal} {\bibinfo  {journal} {J. Mach. Learn. Res.}\ }\textbf {\bibinfo {volume} {24}},\ \bibinfo {pages} {1} (\bibinfo {year} {2023})}\BibitemShut {NoStop}%
\bibitem [{\citenamefont {Louizos}\ \emph {et~al.}(2017)\citenamefont {Louizos}, \citenamefont {Ullrich},\ and\ \citenamefont {Welling}}]{Louizos2017}%
  \BibitemOpen
  \bibfield  {author} {\bibinfo {author} {\bibfnamefont {C.}~\bibnamefont {Louizos}}, \bibinfo {author} {\bibfnamefont {K.}~\bibnamefont {Ullrich}}, \ and\ \bibinfo {author} {\bibfnamefont {M.}~\bibnamefont {Welling}},\ }in\ \href@noop {} {\emph {\bibinfo {booktitle} {Advances in Neural Information Processing Systems (NeurIPS)}}}\ (\bibinfo {year} {2017})\BibitemShut {NoStop}%
\bibitem [{\citenamefont {Louizos}\ \emph {et~al.}(2018)\citenamefont {Louizos}, \citenamefont {Welling},\ and\ \citenamefont {Kingma}}]{Louizos2018}%
  \BibitemOpen
  \bibfield  {author} {\bibinfo {author} {\bibfnamefont {C.}~\bibnamefont {Louizos}}, \bibinfo {author} {\bibfnamefont {M.}~\bibnamefont {Welling}}, \ and\ \bibinfo {author} {\bibfnamefont {D.~P.}\ \bibnamefont {Kingma}},\ }in\ \href@noop {} {\emph {\bibinfo {booktitle} {International Conference on Learning Representations (ICLR)}}}\ (\bibinfo {year} {2018})\BibitemShut {NoStop}%
\end{thebibliography}%

\end{document}